\documentclass[twoside]{article}
\usepackage[preprint]{aistats2026}
\usepackage{colortbl}
\usepackage{hyperref}       
\usepackage{url}            
\usepackage{booktabs}       
\usepackage{amsfonts}       
\usepackage{nicefrac}       
\usepackage{microtype}      
\usepackage{array}          
\usepackage{amsmath}
\usepackage{amsthm}
\newtheorem{proposition}{Proposition}[section] 
\newtheorem{lemma}{Lemma}[section]

\newtheorem*{manualproposition}{Proposition} 
 
\usepackage{algorithm}
\usepackage{algpseudocode} 
\usepackage[table]{xcolor}
\usepackage{multirow}   
\usepackage{graphicx}
\usepackage{courier}
\usepackage[table,xcdraw]{xcolor}      
\usepackage{hyperref}
\usepackage{makecell}
\usepackage{minitoc}
\definecolor{LightIndigo}{HTML}{EDF0FF}
\definecolor{LighGray}{HTML}{D3D3D3}
\usepackage{bm}
\usepackage{etoolbox}
\usepackage{enumitem}
\newcommand{\bc}[1]{\textcolor{black}{#1}}
\usepackage{titletoc}
\usepackage[utf8]{inputenc}
\definecolor{custom_green}{HTML}{00b894}
\definecolor{custom_purple}{HTML}{5f27cd}
\hypersetup{
    colorlinks=true,
    linkcolor=custom_purple,
    filecolor=magenta,      
    urlcolor=custom_green,
    citecolor=custom_green,
}

\DeclareMathOperator*{\argmin}{arg\,min}
\usepackage[round]{natbib}

\begin{document}

%

%
\runningauthor{Islam, Kuipers, Vadgama, Vente, Khan, Sánchez, Bekkers}

\twocolumn[

\aistatstitle{Longitudinal Flow Matching for Trajectory Modeling}
\vspace{-15pt}
\aistatsauthor{%
\mbox{} \And
\textbf{Mohammad Mohaiminul Islam} \And
\mbox{} 
\AND
\textbf{Thijs P. Kuipers} \And
\textbf{Sharvaree Vadgama} \And
\textbf{Coen de Vente}
\AND
\textbf{Afsana Khan} \And
\textbf{Clara I. Sánchez} \And
\textbf{Erik J. Bekkers}
}
\vspace{-60pt}
\aistatsaddress{%
\mbox{} \And
QurAI, UvA \And
\mbox{} 
\AND
BMEP, Amsterdam UMC \And
AMLab, UvA \And
QurAI, UvA
\AND
Maastricht University \And
QurAI, UvA \And
AMLab, UvA 
}
\vspace{-10pt}
{\fontsize{11.25}{14}\selectfont$\quad$\centerline{Open-source code: \href{https://github.com/niazoys/longitudinal-flow-matching}{\texttt{github.com/niazoys/longitudinal-flow-matching}}}
}
\vspace{15pt}
]

\begin{abstract}
Generative models for sequential data often struggle with sparsely sampled and high-dimensional trajectories, typically reducing the learning of dynamics to pairwise transitions. We propose \textit{Interpolative Multi-Marginal Flow Matching} (IMMFM), a framework that learns continuous stochastic dynamics jointly consistent with multiple observed time points. IMMFM employs a piecewise-quadratic interpolation path as a smooth target for flow matching and jointly optimizes drift and a data-driven diffusion coefficient, supported by a theoretical condition for stable learning. This design captures intrinsic stochasticity, handles irregular sparse sampling, and yields subject-specific trajectories. Experiments on synthetic benchmarks and real-world longitudinal neuroimaging datasets show that IMMFM outperforms existing methods in both forecasting accuracy and further downstream tasks.
\end{abstract}

\section{Introduction}

The modeling of trajectories of high-dimensional states is highly relevant in many scientific domains, from climate and geophysical systems \cite{kidger2020neural}, video generation \cite{voleti2021simple, bossa2023multidimensional, dang2023conditional}, and longitudinal biomedical imaging \cite{lachinov2023learning}. Modern continuous-time and generative frameworks have advanced rapidly for modeling these trajectories. Neural ordinary differential equations (NODEs) \citep{chen2018neural} provide a framework for end-to-end learning of ODEs with its time-continuous parametrization. More recently, diffusion based models \citep{shi2023diffusion,liu20232} flow matching \cite{tong2023simulation}, and models based on the Schr{\"o}dinger Bridge approach \cite{hamdouche2023generative} enable the capturing of transitions between arbitrary complex distributions. Neural operator methods also allow for high-resolution data-driven solutions to PDE systems at scale \cite{yang2023fourier}. Together, these developments make it increasingly feasible to model complex and physically consistent trajectories. 


Modeling full high-dimensional trajectories, rather than simplifying them to pairwise transitions \citep{liu2025imageflownet}, remains a key challenge. 
Existing strategies provide partial solutions, but face limitations.
\citet{zhang2024trajectory} introduced a rolling window-based approach, but it lacks temporal alignment and is mainly validated on low-dimensional periodic data.
More broadly, most continuous generative frameworks are formulated as two-marginal transport problems or as pairwise transitions.
Applied naively, these pairwise approaches reduce trajectory learning to a sequence of independent transport, missing constraints, and dependencies across the full trajectory.
Extending these approaches to a multi-marginal setting addresses these limitations \cite{pass2015multi}. 
Concurrent to our work, \citet{lee2025multi} and \citet{rohbeckmodeling} proposed SDE- and ODE-based multi-marginal flow matching methods, respectively, both relying on spline fitting to construct conditional probability paths and thereby addressing the pairwise problem. However, when applied to very high-dimensional data such as images, fitting splines during training can be expensive and unreliable \citep{hastie2009esl, wahba1990spline}, which complicates accurate path learning in these high dimensions.
Thus, practical modeling of sparse, high-dimensional data remains an open challenge.

This challenge is particularly pronounced in clinical medicine, where longitudinal imaging yields repeated views of changing anatomy, yet observations are typically high-dimensional, irregularly and sparsely sampled, and subject-specific.
Accurate modeling of such trajectories could improve treatment selection, streamline follow-up schedules, and support adaptive prognosis and trial design \citep{caruana2015longitudinal,locascio2011overview}.
Traditional pipelines simplify the high-dimensional image data to scalar biomarkers or regional volumes before applying sequence models, discarding rich spatial information \citep{lachinov2023learning,lu2024cats,ruan2024comprehensive,lyu2023multimodal,sun2023manifold,dong2023integrated,liu2020optimizing,lei2020deep,nguyen2023clinically}.
More recent generative approaches operate directly on images but focus primarily on population-level synthesis \citep{wolleb2022medfusion,chen2025generative,zhan2024medm2g,choconditional}, rather than subject-conditioned trajectory modeling needed for prognosis.


To address the challenges of modeling sparse, irregular, and high-dimensional trajectories, we introduce Interpolative Multi-Marginal Flow Matching (IMMFM), designed to capture subject-specific dynamics.IMMFM reframes longitudinal trajectory modeling as a multi-marginal path learning problem. Instead of learning only pairwise transitions, IMMFM jointly learns continuous stochastic dynamics that are consistent with multiple observed time points. We build a piecewise-quadratic conditional interpolation path that supplies a smooth target vector field for flow-matching, and learn both drift and a data-driven diffusion coefficient within a stochastic flow framework. This enables the model to capture intrinsic stochasticity and observation uncertainty. We derive a theoretical condition that supports the joint drift-diffusion learning, ensuring identifiability and stable optimization. Practically, IMMFM respects irregular sampling, enforces temporal smoothness across segments, and yields subject-specific conditional trajectories rather than population synthetic samples.


We summarize our contributions as follows.
\begin{itemize}
    \item We propose a piecewise-quadratic conditional path that yields a smooth and tractable target for flow matching over multiple observations.
    \item We propose to learn a data-driven diffusion coefficient and prove the necessary theoretical condition for joint drift-diffusion optimization.
    \item We demonstrate that IMMFM models trajectories effectively in both a low and high-dimensional setting, showing improved performance over NODE and flow-matching baselines on synthetic benchmarks and longitudinal neuroimaging cohorts.
\end{itemize}

\begin{figure*}
    \centering
    \includegraphics[width=0.85\linewidth]{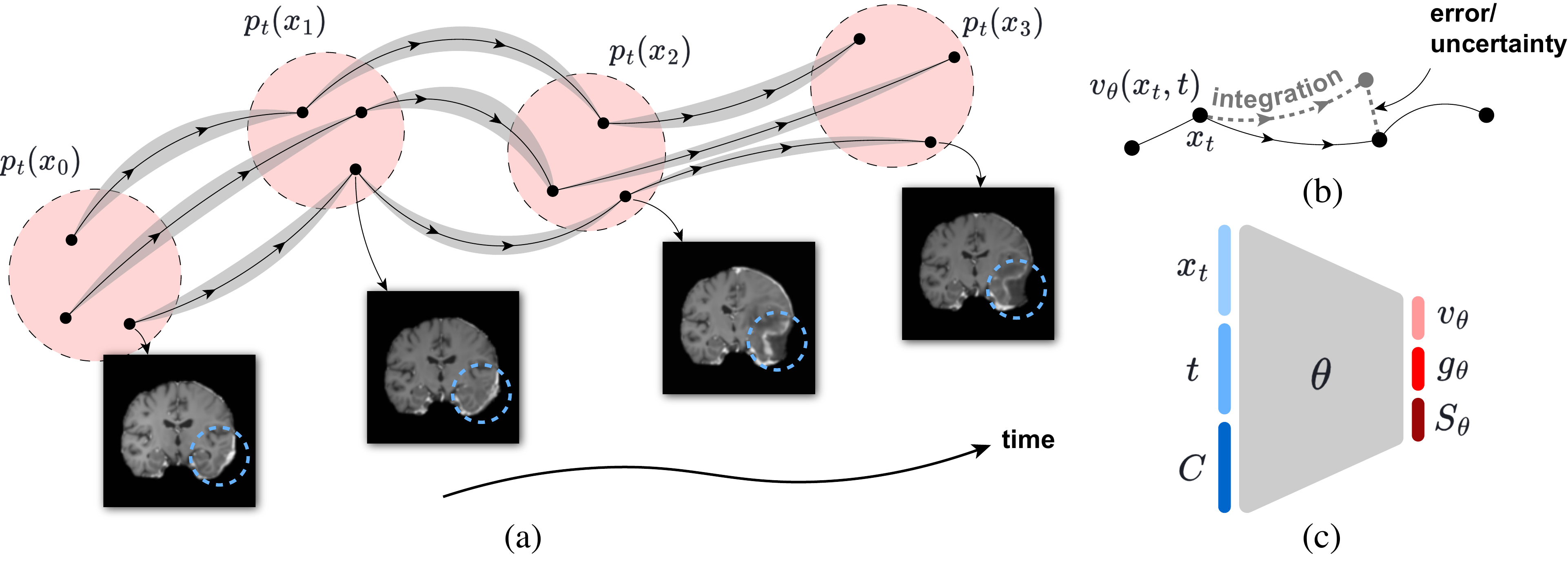}
    \vspace{-10pt}
    \caption{Overview of modeling problem and data. \textbf{(a)} Example of a sparsely and irregularly observed trajectory of disease progression over time. \textbf{(b)}  IMMFM estimates positional uncertainty that informs the SDE's data-driven diffusion term. \textbf{(c)} IMMFM takes as input the position $x_t$, time $t$, and conditional variables $C$ and predicts the velocity $v_\theta$, diffusion term $g_\theta$, and uncertainty $S_\theta$.}
    \label{fig:overview}
\end{figure*}

\section{Background}

Consider \(z_{1:M} = (x_{t_1}, \ldots, x_{t_M})\)  a sequence of observed data or their latent representations acquired at non-uniform and often sparse time points \(t_0 < t_1 < \cdots < t_M \in [0,1]\), with \(x_i \in \mathbb{R}^D\). Let \(\rho_i\) be the probability distribution of the state at time \(t_i\). We assume that these distributions lie on a smooth manifold embedded in \(\mathbb{R}^d\), where \(d \ll D\). The objective is to learn a continuous probabilistic flow \(p_t(x)\) over \(t \in [0,1]\) such that \(p_{t_i} = \rho_i\) for all \(i\) which captures the individual trajectory-specific changes. Let $v(t, x, c)$ be a Lipschitz continuous time-dependent vector field  \(v : [0,1] \times \mathbb{R}^d \times \mathbb{R}^e \rightarrow \mathbb{R}^d\), where \(v(t, x, c)\) is the velocity of the state at time \(t\). The velocity depends on the current position \(x_t\) and conditioning variables \(c \in \mathbb{R}^e\) (e.g. static covariates, baseline measurements, and/or the state at a previous timepoint). The associated flow operator \(\psi_t(v)\) then pushes the initial distribution \(p_0\) forward to \(p_t\). This means that if a sample \(x_0 \sim p_0\), the distribution of the transformed point \(x_t = \psi_t(v)(x_0)\) is \(p_t\).


\subsection{Stochastic Differential Equations (SDEs)}
We consider our trajectory modeling problem as a stochastic process that can be represented as an
\emph{Itô stochastic differential
equation} of the form,
\begin{equation}
d x_t = u_t(x_t)\,dt + g(t,x_t)\,dW_t \, ,
    \label{eq:sde}
\end{equation}
where \(W_t\) is a standard Brownian motion in \(\mathbb{R}^{d}\) and
\(g:[0,1] \times\mathbb{R}^d  \!\to\!\mathbb{R}_+\)
is a diffusion coefficient.
Evolving an initial density \(p_0\) under Eq. ~\eqref{eq:sde} produces a collection of marginal
densities \(\{p_t\}_{t\in[0,1]}\), i.e., the density of $x_t$ at each time $t$, governed by the Fokker--Planck equation:
\begin{equation}
\partial_t p_t(x)
\;=\;
-\,\nabla \cdot \big(p_t\,u_t\big)
\;+\;\frac{1}{2}\,\Delta\!\big(g(t, x_t)^2\,p_t\big).
\label{eq:fokker-planck}
\end{equation}
\bc{where $u_t$ is the drift term, and $g$ is the diffusion coefficient. 
The drift $u_t$ captures the mean progression of trajectories, while the diffusion $g$ quantifies stochastic deviations, providing a general formulation for tracing trajectories across continuous change.
In the deterministic limit when stochastic effects vanish ($g \equiv 0$), this system reduces to an ODE, and the Fokker-Planck equation (Eq.\ref{eq:fokker-planck}) simplifies to the continuity equation of mass transport \citep{gardiner2009stochastic}.}

\subsection{Learning SDEs via Probability Flow and Score Matching}
\label{sec:sde_ode_score_revised}

A fundamental problem in learning SDE (Eq.~\eqref{eq:sde}) dynamics is the high computational demand of conventional training paradigms, often requiring the simulation of numerous full trajectories to estimate gradients \citep[see also][]{kidger2021efficient, ryder2018blackbox, neurips2024_sde_grad, li2020scalable, tzen2019neural}. This simulation process can be prohibitively slow, particularly in high-dimensional settings. An effective alternative is to leverage the fundamental connection between the SDE and its deterministic counterpart: the \emph{probability flow ODE}. This connection provides an insight central to score-based generative modeling and subsequent samplers \citep{song2021score, lu2022dpmsolver, karras2022edm, li2024sharp, cai2025pfode}. We therefore adopt the simulation-free training framework from \citet{tong2023simulationfree} to learn the drift of the probability flow ODE. This ODE is governed by the continuity equation and shares the exact same marginals \(p_t\) as the SDE (Eq.~\ref{eq:sde}):.
\begin{equation}
    \partial_t p_t = -\nabla \cdot (p_t u_t^\circ),
    \label{eq:continuity}
\end{equation}
where \(u_t^\circ\) is the drift of the probability flow. By combining the Fokker-Planck (Eq.~\eqref{eq:fokker-planck}) and continuity (Eq.~\eqref{eq:continuity}) equations, we obtain the following identity (cf.\citep[Eq. 5]{tong2023simulation}):
\begin{equation}
u_t(x_t) = \underbrace{v_t(x_t) + \frac{1}{2}\nabla g(t, x_t)^2}_{\text{Prob. flow drift }u_t^\circ(x_t)} + \frac{g(t,x_t)^2}{2} \nabla \log p_t(x_t).
\label{eq:sde_drift_from_flow_score}
\end{equation}
This identity is key to a simulation-free approach, as it decomposes the SDE drift \(u_t\) into two components that can be learned independently. The first component is the probability flow drift \(u_t^\circ\) that captures the average velocity of the system's evolution\footnote{In our formulation, we \textit{absorb} the term \(\tfrac{1}{2}\nabla(g(t,x)^2)\) into the learnable drift \(u_t^\circ\). This avoids expensive gradient computation w.r.t $g$ during inference while leaving the training objective unchanged.} The second component is the score function \(\nabla \log p_t\) that provides a corrective force to ensure the generated trajectories remain plausible. We utilize this decomposition, and learn the SDE drift by training two separate time-dependent neural networks to approximate both the flow drift \(v_\theta(t, x) \approx u_t^\circ(x)\) and the score function \(s_\theta(t, x) \approx \nabla \log p_t(x)\).
Then the ideal training objective would be to minimize the true marginals \(p_t\): 
\begin{multline}
\mathcal{L}_{\text{SDE}}(\theta) =
\mathbb{E}_{\substack{t \sim \mathcal{U}(0,1)\\ x \sim p_t}}
\big[ \|v_\theta(t, x) - u^\circ(t, x)\|_2^2  \\
+ \lambda(t)^2 \|s_\theta(t, x) - \nabla \log p_t(x)\|_2^2 \big],
\label{eq:flow_score_matching_loss}
\end{multline}
However, this objective is intractable, as the true marginals $p_t$, the probability flow drift $u^\circ(t, x)$, and the score $\nabla \log p_t(x)$ are all unknown.
Instead, we use the tractable conditional objective \citep{tong2023simulation,lipman2022flow}, where targets can be derived analyically for constructed conditional paths \(p_t(x \mid z)\):
\begin{multline}
\mathcal{L}_{\rm CSDE}(\theta)
 = \mathbb{E}_{\substack{t \sim \mathcal{U}(0,1)\\ z \sim q\\ x \sim p_t(x \mid z)}}
 \Biggl[
 \underbrace{\bigl\lVert v_\theta(t,x,c) - u^\circ_t(x \mid z) \bigr\rVert_2^2}_{\text{Cond. Flow Matching}}\\
 +\lambda(t)^2
 \underbrace{\bigl\lVert s_\theta(t,x,c) - \nabla_x \log p_t(x \mid z) \bigr\rVert_2^2}_{\text{Cond. Score Matching}}
 \Biggr].
\label{eq:IMMFM_objective_flow_score}
\end{multline}
The specific analytical forms of target velocity $u^\circ_t(x \mid z)$ and score $\nabla_x \log p_t(x \mid z)$ depend on how the conditional probability path is constructed.

\section{Interpolative Multi-Marginal Flow Matching (IMMFM)}
\label{sec:IMMFM}

As established in \ref{sec:sde_ode_score_revised}, our SDE learning strategy relies on the construction of conditional probability path \( p_t(x \mid z) \) and score  \(\nabla_x \log p_t(x \mid z)\) defined in Eq.~\eqref{eq:IMMFM_objective_flow_score}.

\subsection{Conditional Probability Path Construction} 
Conditional probability paths are commonly defined as Gaussian distributions with time-varying mean and covariance \citep{lipman2022flow}. We follow this approach and define for each trajectory $z \sim q$ the conditional probability path $p_t$:
\begin{equation}
p_t(x\mid z) = \mathcal{N}\!\left(x\bigm|\mu_t(z),\;\sigma^2(t)I\right),
\label{eq:conditional-path}
\end{equation}
where $\mu_t(z): [0,1] \times Z \rightarrow \mathbb{R}^d$ and $\sigma(t): [0,1] \rightarrow \mathbb{R}_+$ are the mean and standard deviation, respectively. Eq~\ref{eq:conditional-path} induces a unique conditional velocity (cf. \citep[Thm.~3]{lipman2022flow}):
\begin{equation}
u^\circ_t(x \mid z) = \frac{\sigma'(t)}{\sigma(t)}(x - \mu_t(z)) + \mu'_t(z).
\label{eq:drift-function}
\end{equation}


In the multi-marginal case, the conditional path is often a simple linear interpolation between the two consecutive points in a trajectory. However, naively chaining these linear paths in a multi-marginal setting leads to a trajectory with discontinuous velocities, which fails to capture the underlying smooth dynamics of the system. We address this by generalizing the path construction, introducing a quadratic term that ensures a smoother transition in velocity. We condition this term on the subsequent trajectory segment to yield probability paths with greater dynamic fidelity. Crucially, this improved consistency is achieved \emph{while preserving the analytical tractability}, which is essential for a simulation-free flow matching objective. We choose $\sigma(t)$ in such a way as to provide the path with a variance structure that is consistent with a Brownian bridge: zero at the observed endpoints ($t_i, t_{i+1}$) and maximum at midway between them. Formally, we define \(\mu_t(z)\) and \(\sigma(t)\) as:
\begin{flalign}
& \mu_t(z) = x_{t_i} + v_i (t - t_i)
  + \tfrac{1}{2}\alpha_t (v_i - v_{i+1})(t - t_i) && \notag\\
& \sigma(t) = \sigma_0 (t - t_i)\,\alpha_t && \label{eq:combined_mu_sigma}
\end{flalign}
where $t$ is strictly between $[t_i,t_{i+1}]$, $v_i=\tfrac{x_{t_{i+1}} - x_{t_i}}{t_{i+1} - t_i}$ is the velocity of the $[t_i, t_{i+1}]$ segment, and $\alpha_t = \frac{t_{i+1}-t}{t_{i+1}-t_i}$ is a time-dependent blending coefficient. This choice of $\mu_t(z)$ and $\sigma(t)$ leads to the associated conditional velocity $u^\circ_t(x|z)$ by substituting their respective derivatives into the general form of Eq.~\eqref{eq:drift-function}. This yields our \emph{blended velocity field}:
\begin{flalign}
& u^\circ_t(x \mid z)
   =v_i + \frac{1}{2}(v_i - v_{i+1})(2\alpha_t - 1) \notag\\
&\qquad \qquad \qquad \qquad  + \frac{\sigma'(t)}{\sigma(t)}\bigl(x - \mu_t(z)\bigr) , &&
\label{eq:blended-drift}
\end{flalign}
where $\mu'_t(z)$ and $\sigma'(t)$ are the time derivatives of $\mu_t(z)$ and $\sigma(t)$, respectively. See Appendix~\ref{mu_sigma_prime} for the full derivation. The corresponding conditional score function that is required for defining the full conditional SDE drift via Eq.~\eqref{eq:sde_drift_from_flow_score}, is analytically tractable for this Gaussian path:
\begin{equation}
\nabla_x \log p_t(x \mid z) = \frac{\mu_t(z)-x}{\sigma^2(t)} \, .
\label{eq:conditional_score}
\end{equation}
These expressions for the blended velocity field in Eq.~\eqref{eq:blended-drift} and the conditional score in Eq.~\eqref{eq:conditional_score} provide the analytically tractable targets for the neural networks $v_\theta$ and $s_\theta$ in our learning objective, Eq.~\eqref{eq:IMMFM_objective_flow_score}.

\begin{figure*}[t]
\centering
\includegraphics[height=200px,width=0.94\linewidth]{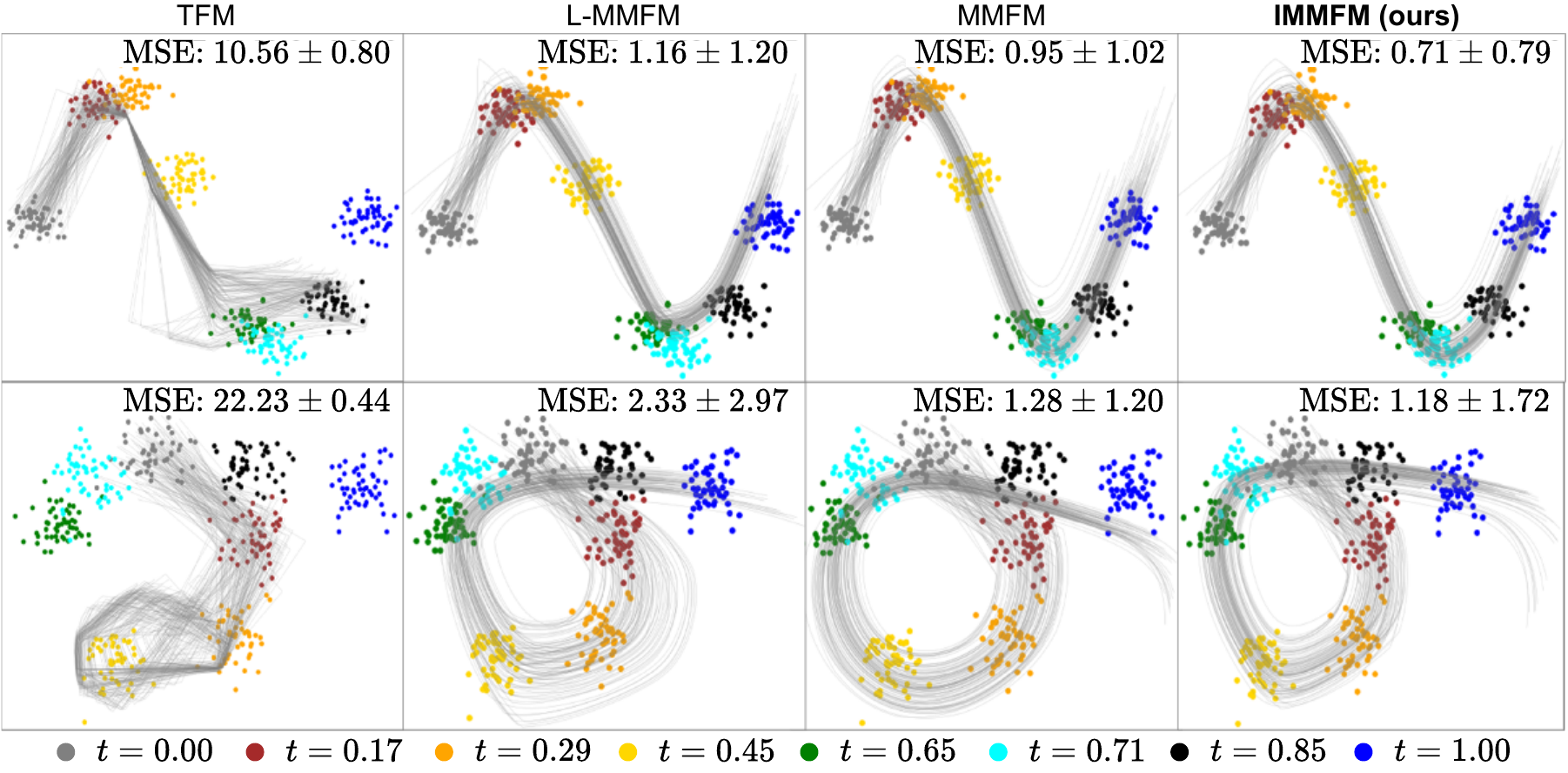}
    \vspace{-12pt}
    \caption{Trajectories on synthetic S-shaped (top row) and $\sigma$-shaped (bottom row) Gaussian datasets. Colored dots show subsets of training samples, and grey lines show the predicted trajectories. From left to right: TFM \citep{zhang2024trajectory}, L-MMFM \citep{rohbeckmodeling}, MMFM \citep{rohbeckmodeling}, and IMMFM (ours).}
    \label{fig:synthetic}
\end{figure*}

\subsection{Uncertainty as a Learned Diffusion Coefficient}
\label{sec:matching_objective}
\bc{
 With the target velocity $u^\circ_t(x \mid z)$ and score $\nabla_x \log p_t(x \mid z)$  of the SDE's drift term now fully specified, the final component of our model is the diffusion coefficient, $g(t)$. We learn the two components of the SDE drift as separate, independent functions. Because these components are learned independently of any specific diffusion schedule, they can be recombined at inference time with an \emph{arbitrary} diffusion coefficient \(g(t)\) to form a valid SDE \citet{tong2023simulation}. We use this flexibility and learn a data-driven diffusion coefficient \(g_\theta\)\footnote{In practice, $g_\theta$ outputs multiple scalars, with Eq.~\ref{eq:sde} and Eq.~\ref{eq:fokker-planck} applied to each.}. More specifically, we let the stochasticity of the model reflect its predictive confidence by matching the diffusion \(g_\theta^2\) to the squared predictive error:}
\begin{multline}
\label{eq:diffusion-position-final}
\mathcal{L}_{\rm uncertainty}(\theta) = \mathbb{E}_{t, z, x} \Biggl[ \Biggl\lVert g_\theta(t,x_t,c)^2 - \\
 \underbrace{\lVert x_i + (t_{i+1}-t_i)u_\theta(t,x_t,c) - x_{t_{i+1}} \rVert_2^2}_{\text{Squared error of predictive construction}} \Biggr\rVert_2^2 \Biggr].
\end{multline}

\subsection{Training Objective}

We formalize the overall training objective of our model, which follows a joint formulation of the primary conditional SDE objective from Eq.~\eqref{eq:IMMFM_objective_flow_score} and the uncertainty objective from Eq.~\eqref{eq:diffusion-position-final}:
\begin{equation}
\mathcal{L}_{\rm IMMFM}(\theta) = \mathcal{L}_{\rm CSDE}(\theta) + \beta  \mathcal{L}_{\rm uncertainty}(\theta),
\label{eq:IMMFM-complete}
\end{equation}
where $\beta$ is a small positive weight. This composite objective is designed for the joint 
optimization of the drift ($v_\theta$), score ($s_\theta$), and diffusion ($g_\theta$) components.

We now establish and prove the theoretical basis of the training objective. 

\begin{proposition}[Gradient equivalence at stationary points]
\label{thm:grad-eq}
Under mild regularity conditions for all $x\in\mathbb R^{d}$, $t\in[0,1]$, every stationary point of $\mathcal L_{\mathrm{SDE}}$ (Eq.~\ref{eq:flow_score_matching_loss}) is a stationary  point of $\mathcal L_{\mathrm{IMMFM}}$ (Eq.~\ref{eq:IMMFM-complete}).
\end{proposition}

The proof relies on the following lemma. 
Here $z$ denotes the trajectory snippet used to construct the conditional path $p_t(x\mid z)$ and the targets (e.g., contains the adjacent observations/timestamps on $[t_i,t_{i+1}]$); its conditional law is $q(z\mid x_t=x, t)$ obtained from the MMOT coupling. 

\begin{lemma}[Zero-mean residual]\label{thm:lemma_zero_mean}
Let $\Delta t := t_{i+1}-t_i$ and define for $t\in(t_i,t_{i+1})$:
\[
r_\theta(t,x,z) \;=\; x_i \;+\; \Delta t\,u_\theta(t,x,c) \;-\; x_{t_{i+1}},
\]
where $u_\theta$ is the assembled drift defined above.
If $u_\theta(t,x,c)=u_t(x)$ then:
\[
\mathbb{E}_{z|x,t}[\,r_\theta(t,x,z)\,]=0.
\]
\begin{proof}
The Markov property of the target SDE gives  
$\mathbb E[x_{t_{i+1}}\mid x_t=x]=x+\Delta t\,u_t(x)$.  
Subtract $x+\Delta t\,u_\theta(t,x,c)$ on both sides and use $u_\theta=u_t$.
\end{proof}
\end{lemma}

With Lemma~\ref{thm:lemma_zero_mean} established, we can now complete the proof of Proposition~\ref{thm:grad-eq}.
\begin{proof} \citet[Thm 3.2]{tong2023simulation} show
$\nabla_{\!\theta}\mathcal L_{\mathrm{C\text{-}SDE}}
=\nabla_{\!\theta}\mathcal L_{\mathrm{SDE}}$ whenever $p_t(x)\!>\!0$.  
Hence, it suffices to prove that
$\nabla_{\!\theta}\mathcal L_{\mathrm{uncertainty}}(\theta^\star)=0$  
at any $\theta^\star$ that minimises $\mathcal L_{\mathrm{SDE}}$. We check whether $\nabla_\theta \mathcal{L}_{\text{uncertainty}}(\theta)$ impacts the optimization problem.
Let $\Delta t:=t_{i+1}-t_i$ and
$r_\theta(t,x,z)=x_i+\Delta t\,u_\theta(t,x,c)-x_{t_{i+1}}$. Then the per-sample uncertainty loss $\ell_{\text{unc}}$ is
\begin{align}
    \ell_{\text{unc}}(\theta; t, x, z) = \big\|g_\theta(t, x, c)^2 - r_\theta(t, x, z)^2\big\|_2^2.
\end{align}
Taking the gradient with respect to $\theta$ and expectation over $z$ conditioned on $(x,t)$:
\begin{multline}    
    \nabla_\theta \ell_{\text{unc}}= 2\big\langle \nabla_\theta g_\theta^2,\, g_\theta^2 - r_\theta^2 \big\rangle \;-\; 2\big\langle g_\theta^2,\, \nabla_\theta r_\theta^2 \big\rangle.
\end{multline}
\begin{multline}
    \mathbb{E}_{z|x,t}[\nabla_\theta \ell_{\text{unc}}] = 2\big\langle \nabla_\theta g_\theta^2,\, g_\theta^2 - \mathbb{E}_{z|x,t}[r_\theta^2] \big\rangle \;-\; \\
    2\big\langle g_\theta^2,\, \mathbb{E}_{z|x,t}[\nabla_\theta r_\theta^2] \big\rangle.
\end{multline}
Under regularity conditions that permit interchange of expectation and differentiation, we have:
\begin{equation}
    \mathbb{E}_{z|x,t}[\nabla_\theta r_\theta^2] = \nabla_\theta \mathbb{E}_{z|x,t}[r_\theta^2] = \nabla_\theta \bar{r}_\theta^2, 
\end{equation}
where $\bar{r}_\theta^2 = \mathbb{E}_{z|x,t}[r_\theta^2]$. When $\theta$ reaches the optimal parameters $\theta^*$ for $\mathcal{L}_{\text{SDE}}$, we assume $v_{\theta^*}(t, x, c) = u_t^{\circ}(x|z)$ and $s_{\theta^*}(t, x, c) = \nabla \log p_t(x|z)$, and $g_{\theta^*}(t,x,c)=g(t,x)$. This implies that $u_{\theta^*}(t, x, c) = u_t(x)$ by the identity introduced earlier, i.e., the learned SDE drift matches the true drift.

By Lemma \ref{thm:lemma_zero_mean}, when $u_{\theta^*} = u_t$, the residual has zero conditional mean, $\mathbb{E}_{z|x,t}[r_{\theta^*}(t, x, z)] = 0$. Consequently, the second moment equals the variance: $\bar{r}_{\theta^*}^2 = \mathbb{E}_{z|x,t}[r_{\theta^*}^2] = \text{Var}_{z|x,t}[r_{\theta^*}]$. As $g_\theta$ is trained to predict this conditional variance,
a standard heteroscedastic-regression objective, we can write $g_{\theta^\star}^{2}=\bar r_{\theta^\star}^{\,2}$ and
$\nabla_{\!\theta}\bar r_{\theta^\star}^{\,2}=0$; both inner products vanish:
\begin{equation}
\mathbb{E}[\nabla_\theta \ell_{\text{unc}}(\theta^*)] = 2\big\langle \nabla_\theta g_{\theta^*}^2,\, 0 \big\rangle \;-\; 2\big\langle g_{\theta^*}^2,\, \nabla_\theta \bar{r}_{\theta^*}^2 \big\rangle \;=\; 0.    
\end{equation}
Thus, we have $\nabla_\theta \mathcal{L}_{\text{uncertainty}}(\theta^*) = 0$, which concludes our proof of Proposition~\ref{thm:grad-eq}.
\end{proof}

\paragraph{Training Objective in Practice.}
The objective in Eq~\eqref{eq:IMMFM-complete} requires complete trajectories for supervision. However, real-world longitudinal datasets rarely provide complete trajectories, but instead consist of sparsely sampled marginal distributions $(\rho_0, \rho_1, \ldots, \rho_M)$ without explicit pairings across time.
We therefore outline a general solution based on multi-marginal optimal transport, which provides a framework to construct the necessary joint distribution $q(z)$ over these sparsely sampled trajectories.

\subsection{Constructing Trajectories via Optimal Transport}
\label{sec:mmot_coupling_unified}
Multi-marginal optimal transport (MMOT) finds the most probable cost-optimal couplings between observations. We formulate MMOT under the assumption of a pairwise additive structure for the global cost \citep{rohbeckmodeling} of the trajectory:
\begin{equation}
C(x_{t_0}, \ldots, x_{t_M}) = \sum_{i=0}^{M-1} k(x_{t_i},x_{t_{i+1}}),
\end{equation}
where $k(\cdot,\cdot)$ represents the transition cost between any two sequential states and $C(\cdots)$ denotes the total cost over the entire trajectory. Under this additive cost structure, the MMOT problem decomposes into a series of independent, simpler pairwise Optimal Transport (OT) problems. The optimal coupling between each consecutive pair of observations $(\rho_i, \rho_{i+1})$ can therefore be found separately. We apply this pairwise OT framework to address the sparsely sampled trajectories of real-world longitudinal datasets. More specifically, the cost $k(x_{t_i}, x_{t_{i+1}})$ becomes the squared Euclidean distance $\|x_{t_i} - x_{t_{i+1}}\|_2^2$, and the minimization is performed over the \emph{augmented empirical distributions} $\rho_i^{\dagger}$ and $\rho_{i+1}^{\dagger}$ \citep{Mok_2020_CVPR}.
These empirical distributions incorporate all smooth diffeomorphic transformations of the trajectory states, effectively turning the MMOT task into a pairwise \textit{spatial alignment} problem to find the optimal transport plan:
\begin{equation}
\pi_{i,i+1}^{*} = \argmin_{\pi \in \Pi(\rho_i^{\dagger}, \rho_{i+1}^{\dagger})} \int_{\mathbb{R}^d \times \mathbb{R}^d} \|x_{t_i} - x_{t_{i+1}}\|_2^2 \, d\pi(x_{t_i},x_{t_{i+1}})
\end{equation}
As a result, the OT framework is a flexible tool to account for any kind of misalignment between trajectory states, given that it can be expressed through a cost function.
After solving for all the optimal pairwise plans $(\pi_{0,1}^{*}, \pi_{1,2}^{*}, \ldots)$ \cite{zhou2024efficient}, they can be combined to reconstruct the full multi-marginal distribution $q(z)$ via the following proposition (proof in Appendix \ref{app:multi-marginal-proof}).
\begin{proposition}[Diffeomorphic MMOT Decomposition]
Under the pairwise additive cost structure, MMOT decomposes into a series of independent pairwise OT problems. The resulting joint coupling is:
\begin{equation}
q(z) = \pi^{*}(x_{t_0}, \ldots, x_{t_M}) = \frac{\prod_{i=0}^{M-1} \pi_{i,i+1}^{*}(x_{t_i}, x_{t_{i+1}})}{\prod_{i=1}^{M-1} \rho_i^{\dagger}(x_{t_i})},
\end{equation}
preserving the intermediate augmented marginals.
\end{proposition}

\begin{figure}[t]
    \centering
    \includegraphics[width=0.98\linewidth]{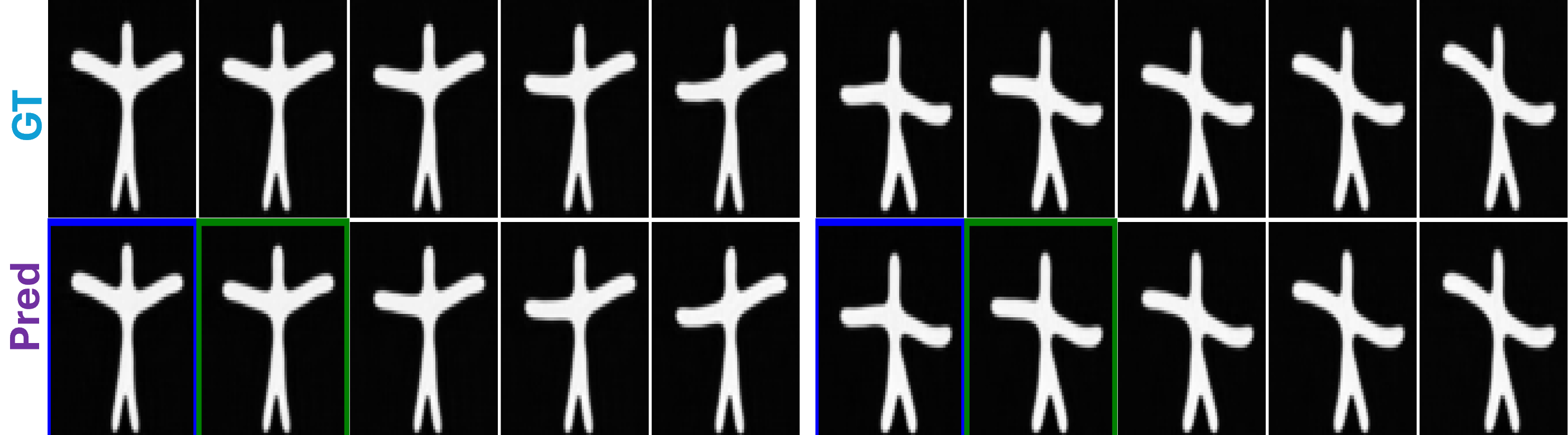}
    \vspace{-12pt}
    \caption{Trajectory on Starmen dataset. The conditioning frame is marked with green, and the reference starting frame is marked with blue. On the left \emph{Hand-downward motion}, on the right \emph{Hand-upward motion}.}
    \label{fig:starmen}
    \end{figure}

\begin{figure*}[t]
    \centering
    \includegraphics[width=0.97\linewidth]{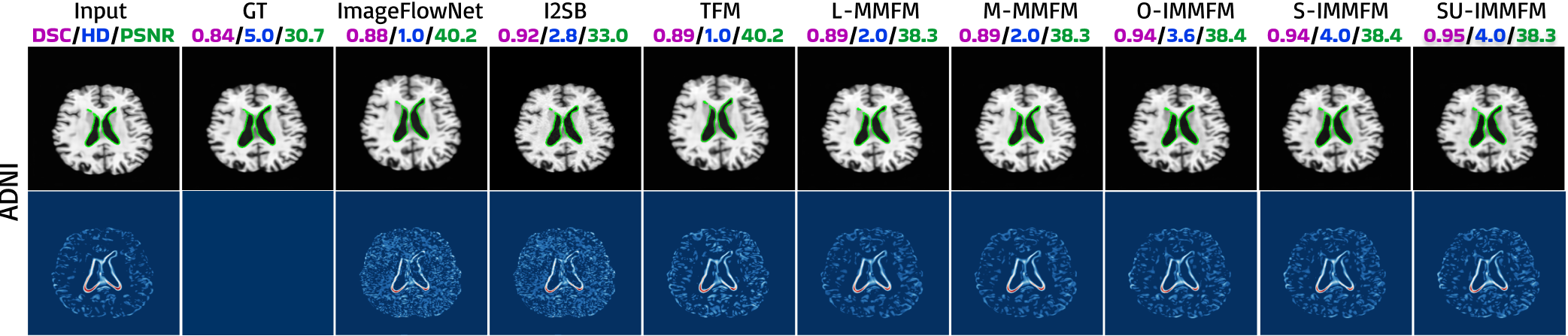}
    \vspace{-12pt}
    \caption{Visual comparison of forecasting results on the Alzheimer's (ADNI), the first row displays the forecasted image. The second row shows the corresponding pixel-wise difference map between the forecast and the ground truth. The different evaluation metrics DSC, HD, and PSNR are listed at the top. }

    \label{fig:all_traj}
\end{figure*}

\begin{figure}[t]
    \centering
    \includegraphics[width=0.95\linewidth]{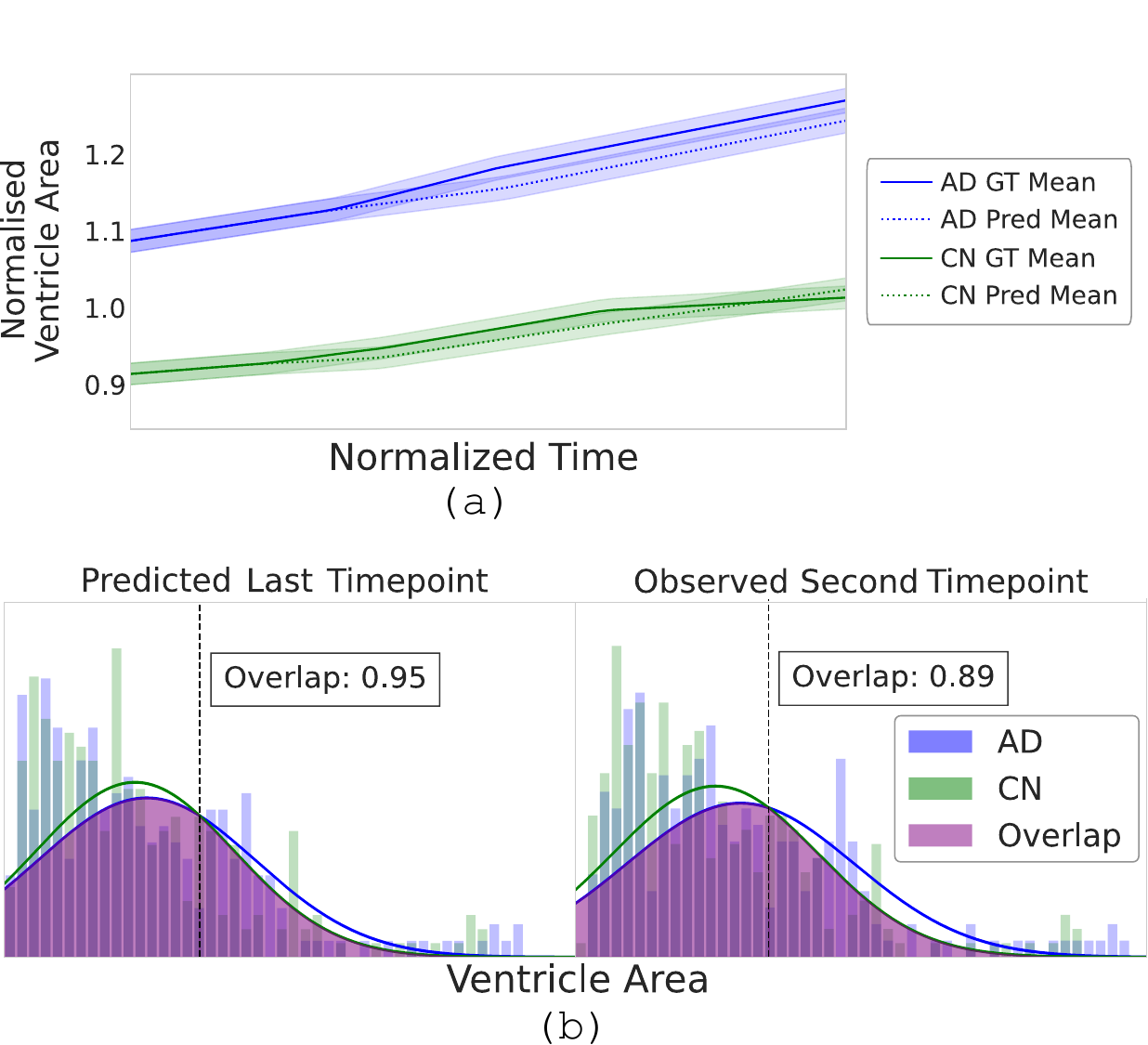}
    \vspace{-12pt}
    \caption{\texttt{(a)} Ground truth and predicted mean ventricle growth over time. \texttt{(b)} Ventricular areas at the second visit ($\sim$18 months) versus model-predicted areas at the last visit ($\sim$36 months) for Alzheimer’s (AD) and cognitively normal (CN) subjects.}
    \label{fig:ad_dist}
\end{figure}

\begin{table*}[t!]
  \centering
  \caption{Trajectory forecasting performance, averaged over three runs and all timepoints. Reported $\pm$ values indicate inter-subject standard deviation.}
  \label{tab:forecasting}
  \small
  \setlength{\tabcolsep}{4.5pt}
  \resizebox{\textwidth}{!}{%
  \begin{tabular}{llcccccccc}
    \toprule
    Dataset & Metric & ImageFlowNet & I2SB & TFM  & L-MMFM & M-MMFM & \textbf{O-IMMFM} & \textbf{S-IMMFM} & \textbf{SU-IMMFM}\\
    \midrule
      Starmen & MSE$\times$10~$\downarrow$ & 0.26$\pm$0.02 & 0.57$\pm$0.37 & 0.23$\pm$0.09  & 0.11$\pm$0.13 & 0.12$\pm$0.14 & \textbf{0.09$\pm$0.01} & 0.09$\pm$0.12 & 0.09$\pm$0.21  \\  
    \midrule
    \multirow{5}{*}{ADNI} 
 & PSNR $\uparrow$ & 36.43$\pm$1.76 & 35.56$\pm$1.97 & 36.01$\pm$4.02 &  36.07$\pm$2.41 & 36.02$\pm$2.45 & 37.51$\pm$2.24 & 37.43$\pm$2.22 & \textbf{37.52$\pm$2.24}\\
  & SSIM $\uparrow$ & 0.94$\pm$0.02 & 0.97$\pm$0.01 & 0.93$\pm$0.11 &  0.96$\pm$0.04 & 0.96$\pm$0.04 & 0.97$\pm$0.06 & 0.97$\pm$0.05  & \textbf{0.97$\pm$0.04} \\
  & MSE$\times$10~$\downarrow$ & 0.04$\pm$0.02 & 0.05$\pm$0.03 & 14.30$\pm$0.03 &  0.02$\pm$0.06 & 0.02$\pm$0.07 & 0.02$\pm$0.07 & \textbf{0.02$\pm$0.05}  & 0.02$\pm$0.06\\
  & DSC  $\uparrow$ & 0.91$\pm$0.15 & 0.91$\pm$0.15 & 0.89$\pm$0.17 & 0.91$\pm$0.19 & 0.9$\pm$0.14 & 0.92$\pm$0.14 & 0.92$\pm$0.14  & \textbf{0.92$\pm$0.14}\\
  & HD   $\downarrow$ & 10.70$\pm$48.27 & 8.78$\pm$46.82 & 11.00$\pm$21.11 &  8.89$\pm$26.74& 11.78$\pm$25.17  & 6.73$\pm$25.01  & 7.17$\pm$29.73 & \textbf{6.50$\pm$22.66}\\
    \midrule
  
\multirow{5}{*}{Brain MS} 
  & PSNR $\uparrow$ & 31.72$\pm$1.83 & 32.77$\pm$0.26 & 31.4$\pm$2.73 & 34.57$\pm$3.24 & 33.75$\pm$3.32 & 36.67$\pm$3.37 & 36.63$\pm$3.28 & \textbf{36.67$\pm$3.36} \\
  & SSIM $\uparrow$ & 0.86$\pm$0.05 & 0.93$\pm$0.03 & 0.88$\pm$0.05 & 0.93$\pm$0.06& 0.92$\pm$0.06 & 0.95$\pm$0.04 & 0.95$\pm$0.04 & \textbf{0.95$\pm$0.04} \\
  & MSE$\times$10~$\downarrow$ & 0.14$\pm$0.10 & 0.05$\pm$0.00 & 13.69$\pm$0.00 & 0.01$\pm$0.01 & 0.02$\pm$0.01 & \textbf{0.01$\pm$0.01}  & \textbf{0.01$\pm$0.01} & \textbf{0.01$\pm$0.01}\\
  & DSC  $\uparrow$ & 0.37$\pm$0.23 & 0.43$\pm$0.22 & 0.51$\pm$0.22 &  0.70$\pm$0.10 & 0.66$\pm$0.11 & 0.73$\pm$0.10 & 0.72$\pm$0.10  & \textbf{0.73$\pm$0.12}\\
  & HD   $\downarrow$ & 63.24$\pm$79.39 & 47.00$\pm$55.39 & 27.71$\pm$0.00  & \textbf{20.57$\pm$10.84} & 24.94$\pm$9.39 & 21.06$\pm$11.17 & 21.17$\pm$11.09 & 21.03$\pm$11.16\\
 \midrule
    \multirow{5}{*}{Brain GBM}
      & PSNR $\uparrow$ & \textbf{35.86$\pm$0.12} & 35.49$\pm$0.17 & 27.47$\pm$12.49 & 30.08$\pm$6.35 & 30.17$\pm$6.86 & 31.78$\pm$5.45 & 31.48$\pm$5.45 & 31.94$\pm$5.55 \\
      & SSIM $\uparrow$ & \textbf{0.94$\pm$0.00} & 0.94$\pm$0.00 & 0.73$\pm$0.37 & 0.89$\pm$0.11  & 0.90$\pm$0.11 & 0.92 $\pm$0.11  & 0.92$\pm$0.21 & 0.93$\pm$0.34 \\
      & MSE$\times$10~$\downarrow$ & \textbf{0.01$\pm$0.01} & 0.02$\pm$0.01 & 0.03$\pm$0.01 & 0.01$\pm$0.01  & 0.01$\pm$0.02 & \textbf{0.01$\pm$0.01}  & \textbf{0.01$\pm$0.01} & \textbf{0.01$\pm$0.01}\\
      & DSC  $\uparrow$ & 0.30$\pm$0.02 & 0.25$\pm$0.00 & 0.34$\pm$0.28 & 0.42$\pm$0.28  & 0.41$\pm$0.28 & 0.45$\pm$0.28 & 0.45$\pm$0.28 & \textbf{0.46$\pm$0.28}\\
      & HD   $\downarrow$ & 198.19$\pm$7.78 & 189.61$\pm$7.64 & 271.5$\pm$15.00 & 142.88$\pm$77.476  & 143.03$\pm$79.12 & 141.36$\pm$77.28 & 141.37$\pm$75.42 & \textbf{135.08$\pm$74.42}\\
    \bottomrule
  \end{tabular}}
\end{table*}

\section{Experimental Setup}
This section details our experimental setup, including the baselines, model implementation, datasets, and evaluation metrics.

\paragraph{Baselines.}
We benchmark our method against recent Neural ODE/SDE, denoising diffusion, and flow matching approaches, including ImageFlowNet \citep{liu2025imageflownet}, $I^2SB$ \citep{liu20232}, and Trajectory Flow Matching (TFM) \citep{zhang2024trajectory}. Two concurrent works proposed related multi-marginal flow matching methods: \citet{lee2025multi}, which combines spline fitting with a rolling-window strategy, and \citet{rohbeckmodeling}, which also relies on spline fitting. Although the former appeared too late for inclusion, we benchmark against the latter (MMFM). However, TFM allows us to evaluate a similar rolling-window strategy. For consistency, we use the same network architecture across all flow matching and diffusion-based methods (see Appendix~\ref{app:architecture}).

\paragraph{Model Implementation.}
IMMFM operates in the latent space obtained from a pre-trained UNet-style autoencoder. The flow matching dynamics are modeled by a U-ViT-based regressor network \citep{davtyan2023efficient}, that learns the components of the SDE. Specifically, we use the neural networks $v_{\theta}(t,x,c)$ and $s_{\theta}(t,x,c)$ to model the drift and score, respectively. We evaluate three variants of IMMFM: 1) a deterministic ODE version (O-IMMFM), where trajectories are generated using Euler integration, 2) a standard SDE version (S-IMMFM) simulated with the Euler-Maruyama method, and 3) an SDE with our learned, uncertainty-driven diffusion coefficient (SU-IMMFM). We incorporate contextual information to generate subject-specific trajectories rather than population averages. The primary component of $c$ is the latent representation of the preceding image $x_{t_{i-1}}$ when modeling the interval $[t_i, t_{i+1}]$. This provides an implicit initial velocity for the trajectory, enabling the learning of relationships between current and previous states to better predict future evolution. Detailed model implementations are provided in Appendix~\ref{app:architecture}.

\paragraph{Datasets} 
We validate our IMMFM variants using several longitudinal datasets, beginning with low-dimensional synthetic benchmarks, including S-shaped and $\sigma$-shaped Gaussians to test the learning of changing curvature and crossover points. Our evaluation also includes the controlled Starmen image dataset \citep{Bone2018} and three real-world clinical cohorts with structural MRI scans of patients with Alzheimer's Disease (ADNI1, 317 participants, ~4-6 visits \citep{Mueller2005}), Multiple Sclerosis (MS) (Brain MS, \bc{19 patients}, ~4-6 visits \citep{Carass2017}), and Glioblastoma (GB) (Brain GBM, 91 patients, ~2-18 visits) \citep{Suter2022}. Each visit represents one time point in the disease progression trajectories of the patients. These clinical datasets present challenges due to their high dimensionality, irregular and sparse sampling, and varying numbers of time points. For all image datasets, we first perform spatial alignment of the images as described in Section \ref{sec:mmot_coupling_unified}, using full volumes for 3D methods and extracted slices for 2D methods. Further details on datasets are provided in Appendix \ref{app:datasets}.

\paragraph{Evaluation Metrics}
We evaluate model performance with 1) image-level similarity metrics and 2) downstream segmentation accuracy metrics. Image quality is assessed with standard synthesis metrics. For pixel-wise error, we use \emph{Mean Squared Error (MSE)} and \emph{Peak Signal-to-Noise Ratio (PSNR)}, a logarithmic measure of reconstruction quality \citep{hore2010image}. To assess the preservation of anatomical features, we also report the \emph{Structural Similarity Index (SSIM)} \citep{wang2004image}. To evaluate the model's ability to forecast clinically relevant changes, we measure the geometric accuracy of key regions of interest (ROI) using baseline metrics to validate medical image segmentation \citep{taha2015metrics,isensee2021nnu,menze2015multimodal,simpson2019msd,maierhein2022metrics,zou2004statistical,crum2006generalized}. The \emph{Dice--S{\o}rensen coefficient (DSC)} measures the volumetric overlap between the predicted and ground-truth ROIs \citep{dice1945measures,sorensen1948method}. The \emph{Hausdorff distance (HD)} complements this by measuring the maximum distance between the boundaries of the two segmentations, quantifying contour accuracy \citep{huttenlocher1993comparing}. For details, see Appendix \ref{app:evaluation_metrics}.

\paragraph{Training and Inference}
The model is trained to learn the components of the SDE. Following \citet{tong2023simulation}, the score weighting $\lambda(t)$ in Eq.~\ref{eq:IMMFM_objective_flow_score} is set to $2\sigma(t)/\sigma_0^2$. At inference, trajectories are simulated by constructing the full SDE drift from Eq.~\eqref{eq:sde_drift_from_flow_score} and the system evolves according to Eq.~\eqref{eq:sde}.
For all experiments, we simulate the trajectory \emph{autoregressively}. Detailed training and inference algorithms are provided in Appendix~\ref{app:algorithms}.

\section{Experiments}

We begin the evaluation of our proposed method by demonstrating trajectory learning performance on a simple multi-marginal dataset of temporally arranged Gaussians representing S-shape and $\sigma$-shape (see Appendix \ref{app:datasets}). We show in Fig. \ref{fig:synthetic} that our method can learn these simple trajectories effectively with improved mean squared error over the baselines. We exclude ImageFlowNet \citep{liu2025imageflownet} and $I^2$SB \citep{liu20232} from this experiment since they were not developed to handle multi-marginal paths.

\subsection{Subject Specific Trajectories via Conditioning}
Starmen have three distinct motion classes: 1) hand-downward motion,
2) hand-upward motion, and 3) static. We train our modelconditioned on the image of the previous time point. As shown in  Fig. \ref{fig:starmen} (and Fig. \ref{fig:starman_large}), our model simulates the trajectory of the correct class solely based on previous frame conditioning. Our method achieves the best MSE compared to all the baselines, see Tab. \ref{tab:forecasting}.

\subsection{Real-World Application: Disease Progression Modeling}
Tab.~\ref{tab:forecasting} summarizes the quantitative results on the disease progression datasets. IMMFM variants consistently outperform baselines across datasets, with gains of $\bm{1}$–$\bm{4.4\%}$ in Dice score, $\bm{1.5}$–$\bm{2.2}$~dB in PSNR, and $\bm{1.2}$–$\bm{4.5\%}$ in SSIM. On noisy, stochastic clinical datasets, the uncertainty-aware SU-IMMFM provides the best performance. We examine the contribution of each model's individual components in an ablation study in Appendix~\ref{app:ablation-study}. We further show in Tab.~\ref{tab:3d-result} that IMMFM extends from 2D image data to 3D volumetric data, improving Dice by $\bm{\sim3\%}$ and reducing inter-subject variability compared to 2D. 

Qualitatively, IMMFM generates consistent images over both short- and long-term horizons (Fig.~\ref{fig:all_traj}, \ref{fig:all_traj_appendix}; see also temporal generalizability in Appendix~\ref{app:temporal-gen}).
The model makes clinically meaningful updates, especially around the lesion and tumor regions, where we observe an improved overlap of these regions with their ground-truth counterparts. 

Finally, we examine the specific progression of the disease. For the ADNI dataset, we focus on the ventricle region, as ventricular enlargement is a well-studied biomarker of AD progression and its underlying neurodegenerative process \citep{nestor2008ventricular}. IMMFM captures distinct ventricular growth trajectories for AD versus CN groups without explicit class conditioning (Fig.~\ref{fig:ad_dist}). Using only observed data at the second visit ($\sim$18 months), AD–CN classification reaches $\bm{71.7\%}$. When using trajectories forecasted by our model to $\sim$36 months, AD-CN classification accuracy increases to $\bm{80.8\%}$, a $\bm{9.1\%}$ gain. In practical terms, this corresponds to correctly classifying about nine additional subjects per 100, \emph{18 months earlier} than would be possible using only the observed data (see Appendix~\ref{app:classification_details} for additional details).

\section{Conclusion}
\label{sec:conclusion}

We introduced Interpolative Multi-Marginal Flow Matching (IMMFM), a framework for learning high-dimensional trajectories from sparse and irregular data. By combining a smooth piecewise-quadratic conditional path with a data-driven diffusion coefficient, IMMFM captures both structured progression and uncertainty. Across synthetic and real datasets, including challenging longitudinal medical imaging benchmarks, IMMFM consistently outperformed prior methods in forecasting accuracy and downstream tasks, demonstrating its improved reliability and clinical relevance. While the approach still depends on high-quality data and sufficient training coverage, future work could address these challenges by developing temporally aware latent spaces \citep{yang2023self}, incorporating biophysical constraints \citep{qian2025physics}, and conditioning on exogenous variables \citep{shaik2024survey}. In conclusion, IMMFM offers a promising foundation for robust, clinically useful trajectory modeling.

%

\bibliography{references}

\begin{thebibliography}{}

\bibitem[Anonymous, 2024]{neurips2024_sde_grad}
Anonymous (2024).
\newblock An efficient high-dimensional gradient estimator for stochastic differential equations.
\newblock In {\em Advances in Neural Information Processing Systems}.
\newblock (NeurIPS 2024, double-blind submission; update with final authors when available).

\bibitem[Avants et~al., 2009]{Avants2009}
Avants, B.~B., Tustison, N.~J., and Song, G. (2009).
\newblock Advanced normalization tools (ants).
\newblock {\em The Insight Journal}, 2(365):1--35.

\bibitem[Blattmann et~al., 2023]{blattmann2023align}
Blattmann, A., Rombach, R., Ling, H., Dockhorn, T., Kim, S.~W., Fidler, S., and Esser, P. (2023).
\newblock Align your latents: High-resolution video synthesis with latent diffusion models.
\newblock In {\em Proceedings of the IEEE/CVF Conference on Computer Vision and Pattern Recognition}, pages 22563--22575.

\bibitem[B{\^o}ne et~al., 2018]{Bone2018}
B{\^o}ne, A., Colliot, O., and Durrleman, S. (2018).
\newblock Learning distributions of shape trajectories from longitudinal datasets: A hierarchical model on a manifold of diffeomorphisms.
\newblock In {\em Proceedings of the IEEE Conference on Computer Vision and Pattern Recognition (CVPR)}, pages 9271--9280.

\bibitem[Bossa and Sahli, 2023]{bossa2023multidimensional}
Bossa, M.~N. and Sahli, H. (2023).
\newblock A multidimensional ode-based model of alzheimer’s disease progression.
\newblock {\em Scientific reports}, 13(1):3162.

\bibitem[Cai et~al., 2025]{cai2025pfode}
Cai, W., Yang, M., and Xie, Y. (2025).
\newblock Minimax optimality of the probability flow {ODE} sampler.
\newblock {\em arXiv preprint arXiv:2501.04567}.

\bibitem[Carass et~al., 2017]{Carass2017}
Carass, A., Roy, S., Jog, A., Cuzzocreo, J.~L., Magrath, E., Gherman, A., Button, J., Nguyen, J., Prados, F., Sudre, C.~H., and et~al. (2017).
\newblock Longitudinal multiple sclerosis lesion segmentation: Resource and challenge.
\newblock {\em NeuroImage}, 148:77--102.

\bibitem[Caruana et~al., 2015]{caruana2015longitudinal}
Caruana, E.~J., Roman, M., Hern{\'a}ndez-S{\'a}nchez, J., and Solli, P. (2015).
\newblock Longitudinal studies.
\newblock {\em Journal of thoracic disease}, 7(11):E537.

\bibitem[Chen et~al., 2018]{chen2018neural}
Chen, R.~T., Rubanova, Y., Bettencourt, J., and Duvenaud, D.~K. (2018).
\newblock Neural ordinary differential equations.
\newblock {\em Advances in neural information processing systems}, 31.

\bibitem[Chen et~al., 2025]{chen2025generative}
Chen, Y. et~al. (2025).
\newblock Generative ai for synthetic data across multiple medical modalities.
\newblock {\em Computers in Biology and Medicine}, 170:107981.

\bibitem[Cho et~al., 2025]{choconditional}
Cho, H., Wei, Z., Lee, S., Dan, T., Wu, G., and Kim, W.~H. (2025).
\newblock Conditional diffusion with ordinal regression: Longitudinal data generation for neurodegenerative disease studies.
\newblock In {\em The Thirteenth International Conference on Learning Representations}.

\bibitem[Crum et~al., 2006]{crum2006generalized}
Crum, W.~R., Camara, O., and Hill, D. L.~G. (2006).
\newblock Generalized overlap measures for evaluation and validation in medical image analysis.
\newblock {\em IEEE Transactions on Medical Imaging}, 25(11):1451--1461.

\bibitem[Cuomo et~al., 2023]{cuomo2023scientific}
Cuomo, S., Di~Cola, V.~S., Giampaolo, F., Rozza, G., Raissi, M., and Piccialli, F. (2023).
\newblock Scientific machine learning.
\newblock {\em Physics Reports}, 1027:1--74.

\bibitem[Dang et~al., 2023]{dang2023conditional}
Dang, T., Han, J., Xia, T., Bondareva, E., Siegele-Brown, C., Chauhan, J., Grammenos, A., Spathis, D., Cicuta, P., and Mascolo, C. (2023).
\newblock Conditional neural ode processes for individual disease progression forecasting: a case study on covid-19.
\newblock In {\em Proceedings of the 29th ACM SIGKDD Conference On Knowledge Discovery and Data Mining}, pages 3914--3925.

\bibitem[Davtyan et~al., 2023]{davtyan2023efficient}
Davtyan, A., Sameni, S., and Favaro, P. (2023).
\newblock Efficient video prediction via sparsely conditioned flow matching.
\newblock In {\em Proceedings of the IEEE/CVF International Conference on Computer Vision}, pages 23263--23274.

\bibitem[Dhariwal and Nichol, 2021]{dhariwal2021diffusion}
Dhariwal, P. and Nichol, A. (2021).
\newblock Diffusion models beat gans on image synthesis.
\newblock {\em Advances in neural information processing systems}, 34:8780--8794.

\bibitem[Dice, 1945]{dice1945measures}
Dice, L.~R. (1945).
\newblock Measures of the amount of ecologic association between species.
\newblock {\em Ecology}, 26(3):297--302.

\bibitem[Dong et~al., 2023]{dong2023integrated}
Dong, X., Wong, R., Lyu, W., Abell-Hart, K., Deng, J., Liu, Y., Hajagos, J.~G., Rosenthal, R.~N., Chen, C., and Wang, F. (2023).
\newblock An integrated lstm-heterorgnn model for interpretable opioid overdose risk prediction.
\newblock {\em Artificial Intelligence in Medicine}, 135:102439.

\bibitem[Dosovitskiy et~al., 2021]{dosovitskiy2021an}
Dosovitskiy, A., Beyer, L., Kolesnikov, A., Weissenborn, D., Zhai, X., Unterthiner, T., Dehghani, M., Minderer, M., Heigold, G., Gelly, S., Uszkoreit, J., and Houlsby, N. (2021).
\newblock An image is worth 16x16 words: Transformers for image recognition at scale.
\newblock In {\em International Conference on Learning Representations (ICLR)}.

\bibitem[Gardiner, 2009]{gardiner2009stochastic}
Gardiner, C.~W. (2009).
\newblock {\em Stochastic Methods: A Handbook for the Natural and Social Sciences}.
\newblock Springer, Berlin, Heidelberg, 5 edition.

\bibitem[Hamdouche et~al., 2023]{hamdouche2023generative}
Hamdouche, M., Henry-Labordere, P., and Pham, H. (2023).
\newblock Generative modeling for time series via schr $\{$$\backslash$" o$\}$ dinger bridge.
\newblock {\em arXiv preprint arXiv:2304.05093}.

\bibitem[Hastie et~al., 2009]{hastie2009esl}
Hastie, T., Tibshirani, R., and Friedman, J. (2009).
\newblock {\em The Elements of Statistical Learning: Data Mining, Inference, and Prediction}.
\newblock Springer Series in Statistics. Springer, New York, 2nd edition.

\bibitem[Hore and Ziou, 2010]{hore2010image}
Hore, A. and Ziou, D. (2010).
\newblock Image quality metrics: A survey.
\newblock In {\em 2010 20th International Conference on Pattern Recognition}, pages 2366--2369. IEEE.

\bibitem[Huttenlocher et~al., 1993]{huttenlocher1993comparing}
Huttenlocher, D.~P., Klanderman, G.~A., and Rucklidge, W.~J. (1993).
\newblock Comparing images using the hausdorff distance.
\newblock {\em IEEE Transactions on Pattern Analysis and Machine Intelligence}, 15(9):850--863.

\bibitem[Inman and Bradley~Jr, 1989]{inman1989}
Inman, H.~F. and Bradley~Jr, E.~L. (1989).
\newblock The overlapping coefficient as a measure of agreement between probability distributions and point estimation of the overlap of two normal densities.
\newblock {\em Communications in Statistics-Theory and Methods}, 18(10):3851--3874.

\bibitem[Isensee et~al., 2021]{isensee2021nnu}
Isensee, F., Jaeger, P.~F., Kohl, S.~A., Petersen, J., and Maier-Hein, K.~H. (2021).
\newblock {nnU-Net}: a self-configuring method for deep learning-based biomedical image segmentation.
\newblock {\em Nature Methods}, 18(2):203--211.

\bibitem[Karras et~al., 2022]{karras2022edm}
Karras, T., Aittala, M., Laine, S., Herva, A., and Lehtinen, J. (2022).
\newblock Elucidating the design space of diffusion-based generative models.
\newblock {\em arXiv preprint arXiv:2206.00364}.

\bibitem[Kidger et~al., 2021]{kidger2021efficient}
Kidger, P., Foster, J., Li, R. T.~Q., and Lyons, T. (2021).
\newblock Efficient and accurate gradients for neural {SDE}s.
\newblock In {\em International Conference on Machine Learning}, pages 5562--5572.

\bibitem[Kidger et~al., 2020]{kidger2020neural}
Kidger, P., Morrill, J., Foster, J., and Lyons, T. (2020).
\newblock Neural controlled differential equations for irregular time series.
\newblock {\em Advances in neural information processing systems}, 33:6696--6707.

\bibitem[Lachinov et~al., 2023]{lachinov2023learning}
Lachinov, D., Chakravarty, A., Grechenig, C., Schmidt-Erfurth, U., and Bogunovi{\'c}, H. (2023).
\newblock Learning spatio-temporal model of disease progression with neuralodes from longitudinal volumetric data.
\newblock {\em IEEE Transactions on Medical Imaging}, 43(3):1165--1179.

\bibitem[Lee et~al., 2025]{lee2025multi}
Lee, J., Moradijamei, B., and Shakeri, H. (2025).
\newblock Multi-marginal stochastic flow matching for high-dimensional snapshot data at irregular time points.
\newblock {\em arXiv preprint arXiv:2508.04351}.

\bibitem[Lei et~al., 2020]{lei2020deep}
Lei, B., Yang, M., Yang, P., Zhou, F., Hou, W., Zou, W., Li, X., Wang, T., Xiao, X., and Wang, S. (2020).
\newblock Deep and joint learning of longitudinal data for alzheimer’s disease prediction.
\newblock {\em Pattern Recognition}, 102:107247.

\bibitem[Li et~al., 2024]{li2024sharp}
Li, G., Zhang, Y., and Cheng, X. (2024).
\newblock A sharp convergence theory for the probability flow {ODE}.
\newblock {\em arXiv preprint arXiv:2404.01234}.

\bibitem[Li et~al., 2020]{li2020scalable}
Li, R. T.~Q., Kidger, P., Foster, J., and Lyons, T. (2020).
\newblock Scalable gradients for stochastic differential equations.
\newblock In {\em Advances in Neural Information Processing Systems}, volume~33, pages 3515--3527.

\bibitem[Lipman et~al., 2022]{lipman2022flow}
Lipman, Y., Chen, R.~T., Ben-Hamu, H., Nickel, M., and Le, M. (2022).
\newblock Flow matching for generative modeling.
\newblock {\em arXiv preprint arXiv:2210.02747}.

\bibitem[Liu et~al., 2025]{liu2025imageflownet}
Liu, C., Xu, K., Shen, L.~L., Huguet, G., Wang, Z., Tong, A., Bzdok, D., Stewart, J., Wang, J.~C., Del~Priore, L.~V., et~al. (2025).
\newblock Imageflownet: Forecasting multiscale image-level trajectories of disease progression with irregularly-sampled longitudinal medical images.
\newblock In {\em ICASSP 2025-2025 IEEE International Conference on Acoustics, Speech and Signal Processing (ICASSP)}, pages 1--5. IEEE.

\bibitem[Liu et~al., 2023]{liu20232}
Liu, G.-H., Vahdat, A., Huang, D.-A., Theodorou, E.~A., Nie, W., and Anandkumar, A. (2023).
\newblock I\textsuperscript{2}sb: Image-to-image schr{\"o}dinger bridge.
\newblock {\em arXiv preprint arXiv:2302.05872}.

\bibitem[Liu et~al., 2020]{liu2020optimizing}
Liu, P., Fu, B., Yang, S.~X., Deng, L., Zhong, X., and Zheng, H. (2020).
\newblock Optimizing survival analysis of xgboost for ties to predict disease progression of breast cancer.
\newblock {\em IEEE Transactions on Biomedical Engineering}, 68(1):148--160.

\bibitem[Locascio and Atri, 2011]{locascio2011overview}
Locascio, J.~J. and Atri, A. (2011).
\newblock An overview of longitudinal data analysis methods for neurological research.
\newblock {\em Dementia and geriatric cognitive disorders extra}, 1(1):330--357.

\bibitem[Lu et~al., 2022]{lu2022dpmsolver}
Lu, C., Zhou, Z., Bao, F., Chen, J., Li, C., and Zhu, J. (2022).
\newblock {DPM}-solver: A fast {ODE} solver for diffusion probabilistic model sampling in around 10 steps.
\newblock {\em Advances in Neural Information Processing Systems}, 35:5775--5788.

\bibitem[Lu et~al., 2024]{lu2024cats}
Lu, J., Han, X., Sun, Y., and Yang, S. (2024).
\newblock Cats: Enhancing multivariate time series forecasting by constructing auxiliary time series as exogenous variables.
\newblock In {\em Proceedings of the Forty-first International Conference on Machine Learning (ICML)}.
\newblock arXiv preprint arXiv:2403.01673.

\bibitem[Lyu et~al., 2023]{lyu2023multimodal}
Lyu, W., Dong, X., Wong, R., Zheng, S., Abell-Hart, K., Wang, F., and Chen, C. (2023).
\newblock A multimodal transformer: Fusing clinical notes with structured ehr data for interpretable in-hospital mortality prediction.
\newblock In {\em AMIA Annual Symposium Proceedings}, volume 2022, page 719.

\bibitem[Maier-Hein et~al., 2022]{maierhein2022metrics}
Maier-Hein, L., Reinke, A., et~al. (2022).
\newblock Metrics reloaded: recommendations for image analysis validation.
\newblock {\em Nature Methods}, 19(9):1127--1140.

\bibitem[Menze et~al., 2015]{menze2015multimodal}
Menze, B.~H., Jakab, A., Bauer, S., Kalpathy-Cramer, J., Farahani, K., Kirby, J., Burren, Y., Porz, N., Slotboom, J., Wiest, R., et~al. (2015).
\newblock The multimodal brain tumor image segmentation benchmark (brats).
\newblock In {\em IEEE Transactions on Medical Imaging}, volume~34, pages 1993--2024.

\bibitem[Mok and Chung, 2020]{Mok_2020_CVPR}
Mok, T.~C. and Chung, A.~C. (2020).
\newblock Fast symmetric diffeomorphic image registration with convolutional neural networks.
\newblock In {\em Proceedings of the IEEE/CVF Conference on Computer Vision and Pattern Recognition (CVPR)}.

\bibitem[Mueller et~al., 2005]{Mueller2005}
Mueller, S.~G., Weiner, M.~W., Thal, L.~J., Petersen, R.~C., Jack, C.~R., Jagust, W., Trojanowski, J.~Q., Toga, A.~W., and Beckett, L. (2005).
\newblock Ways toward an early diagnosis in {Alzheimer's} disease: {The Alzheimer's Disease Neuroimaging Initiative} {(ADNI)}.
\newblock {\em Alzheimer's \& Dementia}, 1(1):55--66.

\bibitem[Nestor et~al., 2008]{nestor2008ventricular}
Nestor, S.~M., Rupsingh, R., Borrie, M., Smith, M., Accomazzi, V., Wells, J.~L., Fogarty, J., Bartha, R., and Initiative, A. D.~N. (2008).
\newblock Ventricular enlargement as a possible measure of alzheimer's disease progression validated using the alzheimer's disease neuroimaging initiative database.
\newblock {\em Brain}, 131(9):2443--2454.

\bibitem[Nguyen et~al., 2023]{nguyen2023clinically}
Nguyen, H.~H., Blaschko, M.~B., Saarakkala, S., and Tiulpin, A. (2023).
\newblock Clinically inspired multi-agent transformers for disease trajectory forecasting from multimodal data.
\newblock {\em IEEE Transactions on Medical Imaging}.

\bibitem[Pass, 2015]{pass2015multi}
Pass, B. (2015).
\newblock Multi-marginal optimal transport: theory and applications.
\newblock {\em ESAIM: Mathematical Modelling and Numerical Analysis}, 49(6):1771--1790.

\bibitem[Qian et~al., 2025]{qian2025physics}
Qian, Y., Marty, {\'E}., Basu, A., O’Dea, E.~B., Wang, X., Fox, S., Rohani, P., Drake, J.~M., and Li, H. (2025).
\newblock Physics-informed deep learning for infectious disease forecasting.
\newblock {\em ArXiv}, pages arXiv--2501.

\bibitem[Reiser and Faraggi, 1999]{reiser1999}
Reiser, B. and Faraggi, D. (1999).
\newblock Measuring the effectiveness of diagnostic markers in the presence of measurement error through the use of roc curves.
\newblock {\em Statistics in medicine}, 18(17-18):2583--2600.

\bibitem[Rohbeck et~al., 2025]{rohbeckmodeling}
Rohbeck, M., De~Brouwer, E., Bunne, C., Huetter, J.-C., Biton, A., Chen, K.~Y., Regev, A., and Lopez, R. (2025).
\newblock Modeling complex system dynamics with flow matching across time and conditions.
\newblock In {\em The Thirteenth International Conference on Learning Representations}.

\bibitem[Ruan et~al., 2024]{ruan2024comprehensive}
Ruan, C., Huang, C., and Yang, Y. (2024).
\newblock Comprehensive evaluation of multimodal ai models in medical imaging diagnosis: From data augmentation to preference-based comparison.
\newblock {\em arXiv preprint arXiv:2412.05536}.

\bibitem[Ryder et~al., 2018]{ryder2018blackbox}
Ryder, T., Dondelinger, F., and Husmeier, D. (2018).
\newblock Black-box variational inference for stochastic differential equations.
\newblock In {\em International Conference on Artificial Intelligence and Statistics}, pages 1281--1290.

\bibitem[Shaik et~al., 2024]{shaik2024survey}
Shaik, T., Tao, X., Li, L., Xie, H., and Vel{\'a}squez, J.~D. (2024).
\newblock A survey of multimodal information fusion for smart healthcare: Mapping the journey from data to wisdom.
\newblock {\em Information Fusion}, 102:102040.

\bibitem[Shi et~al., 2023]{shi2023diffusion}
Shi, Y., De~Bortoli, V., Campbell, A., and Doucet, A. (2023).
\newblock Diffusion schr{\"o}dinger bridge matching.
\newblock {\em Advances in Neural Information Processing Systems}, 36:62183--62223.

\bibitem[Simpson et~al., 2019]{simpson2019msd}
Simpson, A.~L., Antonelli, M., Bakas, S., Bilello, M., Farahani, K., Van~Ginneken, B., Kopp-Schneider, A., Landman, B.~A., Litjens, G., Menze, B.~H., et~al. (2019).
\newblock A large annotated medical image dataset for the development and evaluation of segmentation algorithms.
\newblock {\em arXiv preprint arXiv:1902.09063}.

\bibitem[Song et~al., 2021]{song2021score}
Song, Y., Sohl-Dickstein, J., Kingma, D.~P., Kumar, A., Ermon, S., and Poole, B. (2021).
\newblock Score-based generative modeling through stochastic differential equations.
\newblock In {\em International Conference on Learning Representations}.

\bibitem[S{\o}rensen, 1948]{sorensen1948method}
S{\o}rensen, T. (1948).
\newblock A method of establishing groups of equal amplitude in plant sociology based on similarity of species and its application to analyses of the vegetation on danish commons.
\newblock {\em Biologiske Skrifter}, 5:1--34.

\bibitem[Sun and Yang, 2023]{sun2023manifold}
Sun, Y. and Yang, S. (2023).
\newblock Manifold-constrained gaussian process inference for time-varying parameters in dynamic systems.
\newblock {\em Statistics and Computing}, 33(6):142.

\bibitem[Suter et~al., 2022]{Suter2022}
Suter, Y., Knecht, U., Valenzuela, W., Notter, M., Hewer, E., Schucht, P., Wiest, R., and Reyes, M. (2022).
\newblock The lumiere dataset: Longitudinal glioblastoma mri with expert rano evaluation.
\newblock {\em Scientific Data}, 9(1):768.

\bibitem[Taha and Hanbury, 2015]{taha2015metrics}
Taha, A.~A. and Hanbury, A. (2015).
\newblock Metrics for evaluating 3d medical image segmentation: analysis, selection, and tool.
\newblock {\em BMC Medical Imaging}, 15(1):29.

\bibitem[Tong et~al., 2023]{tong2023simulation}
Tong, A., Malkin, N., Fatras, K., Atanackovic, L., Zhang, Y., Huguet, G., Wolf, G., and Bengio, Y. (2023).
\newblock Simulation-free schr{\"o}dinger bridges via score and flow matching.
\newblock {\em arXiv preprint arXiv:2307.03672}.

\bibitem[Tong et~al., 2024]{tong2023simulationfree}
Tong, A., Shi, Y., Maddison, C.~J., and Amos, B. (2024).
\newblock Simulation-free {Schr{\"o}dinger} bridges via score and flow matching.
\newblock In {\em Proceedings of the International Conference on Machine Learning}.

\bibitem[Tzen and Raginsky, 2019]{tzen2019neural}
Tzen, B. and Raginsky, M. (2019).
\newblock Neural stochastic differential equations: Deep latent gaussian models in continuous time.
\newblock In {\em International Conference on Machine Learning}, pages 5474--5483.

\bibitem[Vaswani et~al., 2017]{vaswani2017attention}
Vaswani, A., Shazeer, N., Parmar, N., Uszkoreit, J., Jones, L., Gomez, A.~N., Kaiser, {\L}., and Polosukhin, I. (2017).
\newblock Attention is all you need.
\newblock In {\em Advances in Neural Information Processing Systems}, pages 5998--6008.

\bibitem[Voleti et~al., 2021]{voleti2021simple}
Voleti, V., Kanaa, D., Kahou, S.~E., and Pal, C. (2021).
\newblock Simple video generation using neural odes.
\newblock {\em arXiv preprint arXiv:2109.03292}.

\bibitem[von K{\"u}gelgen et~al., 2024]{von2024causal}
von K{\"u}gelgen, J., Gresele, L., and Sch{\"o}lkopf, B. (2024).
\newblock Causal representation learning.
\newblock {\em arXiv preprint arXiv:2401.14411}.

\bibitem[Wahba, 1990]{wahba1990spline}
Wahba, G. (1990).
\newblock {\em Spline Models for Observational Data}, volume~59 of {\em CBMS–NSF Regional Conference Series in Applied Mathematics}.
\newblock SIAM, Philadelphia, PA.

\bibitem[Wang et~al., 2004]{wang2004image}
Wang, Z., Bovik, A.~C., Sheikh, H.~R., and Simoncelli, E.~P. (2004).
\newblock Image quality assessment: From error visibility to structural similarity.
\newblock {\em IEEE Transactions on Image Processing}, 13(4):600--612.

\bibitem[Wolleb et~al., 2022]{wolleb2022medfusion}
Wolleb, J. et~al. (2022).
\newblock Medfusion: Diffusion models for medical image synthesis.
\newblock {\em Medical Image Analysis}, 79:102479.

\bibitem[Wu et~al., 2024]{wu2024q}
Wu, Y., Ku, Z., Yin, P., and Zhang, Y. (2024).
\newblock Q-fusion: A modular framework for multimodal multi-task foundation models.
\newblock {\em arXiv preprint arXiv:2402.13220}.

\bibitem[Yang et~al., 2023a]{yang2023self}
Yang, D., Wang, Y., Kong, Q., Dantcheva, A., Garattoni, L., Francesca, G., and Br{\'e}mond, F. (2023a).
\newblock Self-supervised video representation learning via latent time navigation.
\newblock In {\em Proceedings of the AAAI Conference on Artificial Intelligence}, volume~37, pages 3118--3126.

\bibitem[Yang et~al., 2023b]{yang2023fourier}
Yang, Q., Hernandez-Garcia, A., Harder, P., Ramesh, V., Sattegeri, P., Szwarcman, D., and Rolnick, D. (2023b).
\newblock Fourier neural operators for arbitrary resolution climate data downscaling. arxiv.
\newblock {\em arXiv preprint arXiv:2305.14452}, 10.

\bibitem[Yu et~al., 2022]{yu2022unified}
Yu, S., Chen, T., Shen, J., Yuan, H., Tan, J., Yang, S., Liu, J., and Wang, Z. (2022).
\newblock Unified visual transformer compression.
\newblock In {\em Proceedings of the International Conference on Learning Representations (ICLR)}.

\bibitem[Zhan et~al., 2024]{zhan2024medm2g}
Zhan, Y. et~al. (2024).
\newblock Medm2g: Unifying medical multi-modal generation via cross-guided diffusion with visual invariance.
\newblock In {\em Proceedings of the IEEE/CVF Conference on Computer Vision and Pattern Recognition (CVPR)}.

\bibitem[Zhang et~al., 2024]{zhang2024trajectory}
Zhang, X., Pu, Y., Kawamura, Y., Loza, A., Bengio, Y., Shung, D.~L., and Tong, A. (2024).
\newblock Trajectory flow matching with applications to clinical time series modeling.
\newblock {\em arXiv preprint arXiv:2410.21154}.

\bibitem[Zhou and Parno, 2024]{zhou2024efficient}
Zhou, B. and Parno, M. (2024).
\newblock Efficient and exact multimarginal optimal transport with pairwise costs.
\newblock {\em Journal of Scientific Computing}, 100(1):25.

\bibitem[Zou et~al., 2004]{zou2004statistical}
Zou, K.~H., Warfield, S.~K., Bharatha, A., Tempany, C. M.~C., Kaus, M.~R., Haker, S.~J., Wells, W.~M., Jolesz, F.~A., and Kikinis, R. (2004).
\newblock Statistical validation of image segmentation quality based on a spatial overlap index: scientific reports.
\newblock {\em Academic Radiology}, 11(2):178--189.

\end{thebibliography}

\clearpage
\appendix
\thispagestyle{empty}

\onecolumn
\aistatstitle{Longitudinal Flow Matching For Trajectory Modeling: \\
Supplementary Materials}
\vspace{-40pt}
\startcontents[appendices]
\printcontents[appendices]{}{1}{\section*{}\setcounter{tocdepth}{1}}
\rule{\linewidth}{0.6pt}

\section{Proof of Theoretical Results}
\label{app:theory_proof}

\begin{manualproposition}[\textbf{3.2}]
Under the pairwise additive cost structure, the MMOT problem decomposes into a series of independent pairwise OT problems. The resulting joint coupling $q(z)$ is:
\begin{equation}
q(z) = \pi^{*}(x_{t_0}, \ldots, x_{t_M}) = \frac{\prod_{i=0}^{M-1} \pi_{i,i+1}^{*}(x_{t_i}, x_{t_{i+1}})}{\prod_{i=1}^{M-1} \rho_i^{\dagger}(x_{t_i})}
\end{equation}
This coupling preserves the intermediate augmented marginals.
\label{app:multi-marginal-proof}
\begin{proof}
We consider the multi-marginal optimal transport problem with augmented distributions $\rho_i^{\dagger}$ that incorporate all possible diffeomorphic transformations:
\begin{equation}
\min_{\pi \in \Pi(\rho_0^{\dagger}, \ldots, \rho_M^{\dagger})} \int C(x_{t_0}, \ldots, x_{t_M}) \, d\pi(x_{t_0}, \ldots, x_{t_M})
\end{equation}
where $\Pi(\rho_0^{\dagger}, \ldots, \rho_M^{\dagger})$ denotes couplings with prescribed augmented marginals.

Using the pairwise additive structure, the objective becomes:
\begin{align}
\min_{\pi \in \Pi(\rho_0^{\dagger}, \ldots, \rho_M^{\dagger})} \int \sum_{i=0}^{M-1} c(x_{t_i}, x_{t_{i+1}}) \, d\pi(x_{t_0}, \ldots, x_{t_M})
= \min_{\pi \in \Pi(\rho_0^{\dagger}, \ldots, \rho_M^{\dagger})} \sum_{i=0}^{M-1} \int c(x_{t_i}, x_{t_{i+1}}) \, d\pi_{i,i+1}(x_{t_i}, x_{t_{i+1}})
\end{align}
where $\pi_{i,i+1}$ is the marginal of $\pi$ on coordinates $(x_{t_i}, x_{t_{i+1}})$.

We claim that the coupling:
\begin{equation}
q(z) = \frac{\prod_{i=0}^{M-1} \pi_{i,i+1}^{*}(x_{t_i}, x_{t_{i+1}})}{\prod_{i=1}^{M-1} \rho_i^{\dagger}(x_{t_i})}
\end{equation}
solves the multi-marginal problem. To verify this, we must show it preserves marginals and minimizes cost.

For marginal preservation, we verify that $q(z)$ has the correct marginals. For interior points $i \in \{1,\ldots,M-1\}$:
\begin{align}
\int q(z) \prod_{j \neq i} dx_{t_j} &= \int \frac{\prod_{k=0}^{M-1} \pi_{k,k+1}^{*}(x_{t_k},x_{t_{k+1}})}{\prod_{j=1}^{M-1}\rho_j^{\dagger}(x_{t_j})} \prod_{j \neq i} dx_{t_j}\\
&= \frac{1}{\rho_i^{\dagger}(x_{t_i})} \int \pi_{i-1,i}^{*}(x_{t_{i-1}},x_{t_i}) \pi_{i,i+1}^{*}(x_{t_i},x_{t_{i+1}}) \prod_{k \notin \{i-1,i,i+1\}} \pi_{k,k+1}^{*}(x_{t_k},x_{t_{k+1}}) \prod_{j \neq i} dx_{t_j}
\end{align}

Using Fubini's theorem to integrate out variables that appear in only one factor:
\begin{align}
&= \frac{1}{\rho_i^{\dagger}(x_{t_i})} \left[\int \pi_{i-1,i}^{*}(x_{t_{i-1}},x_{t_i}) \, dx_{t_{i-1}}\right] \left[\int \pi_{i,i+1}^{*}(x_{t_i},x_{t_{i+1}}) \, dx_{t_{i+1}}\right]\\
&= \frac{1}{\rho_i^{\dagger}(x_{t_i})} \cdot \rho_i^{\dagger}(x_{t_i}) \cdot \rho_i^{\dagger}(x_{t_i}) = \rho_i^{\dagger}(x_{t_i})
\end{align}

For boundary points:
\begin{align}
\int q(z) \prod_{j \neq 0} dx_{t_j} &= \int \pi_{0,1}^{*}(x_{t_0},x_{t_1}) \, dx_{t_1} = \rho_0^{\dagger}(x_{t_0})\\
\int q(z) \prod_{j \neq M} dx_{t_j} &= \int \pi_{M-1,M}^{*}(x_{t_{M-1}},x_{t_M}) \, dx_{t_{M-1}} = \rho_M^{\dagger}(x_{t_M})
\end{align}

For cost minimization, the cost under $q(z)$ is
\[
\int C(z)\,dq(z)
= \sum_{i=0}^{M-1}
  \int c\bigl(x_{t_i},x_{t_{i+1}}\bigr)\,
     d\pi_{i,i+1}^{*}\bigl(x_{t_i},x_{t_{i+1}}\bigr),
\]
which is the sum of optimal pairwise costs. Since each $\pi_{i,i+1}^{*}$ minimizes its respective term, $q(z)$ minimizes the total cost.
\end{proof}
\end{manualproposition}

\subsection{Derivative of Mean $\mu_t$ and Varinace $\sigma(t)$} \label{mu_sigma_prime}
We formulate our \emph{blended velocity field} as:
\begin{align}
& u^\circ_t(x \mid z)
   =v_i + \frac{1}{2}(v_i - v_{i+1})(2\alpha_t - 1) + \frac{\sigma'(t)}{\sigma(t)}\bigl(x - \mu_t(z)\bigr) , &&
\label{app:blended-drift}
\end{align}
where $\mu'_t(z)$ and $\sigma'(t)$ are the time derivatives of $\mu_t(z)$ and $\sigma(t)$, respectively.
The mean function for this case is:
\begin{align}
\mu_t &= x_i + \frac{x_{i+1} - x_i}{t_{i+1} - t_i}(t - t_i) + \frac{1}{2}\left(\frac{x_{i+1} - x_i}{t_{i+1} - t_i} - \frac{x_{i+2} - x_{i+1}}{t_{i+2} - t_{i+1}}\right)\frac{(t - t_i)(t_{i+1} - t)}{t_{i+1} - t_i} \\
&= x_i + v_i(t - t_i) + \frac{1}{2}(v_i - v_{i+1})\frac{(t - t_i)(t_{i+1} - t)}{t_{i+1} - t_i}
\end{align}

where the scaled segment velocities are:
\begin{align}
v_i &= \frac{x_{i+1} - x_i}{t_{i+1} - t_i} \\
v_{i+1} &= \frac{x_{i+2} - x_{i+1}}{t_{i+2} - t_{i+1}}
\end{align}

Derivative of $\mu'_t$ function:
\begin{align}
\mu'_t &= \frac{d}{dt}\left[x_i + v_i(t - t_i) + \frac{1}{2}(v_i - v_{i+1})\frac{(t - t_i)(t_{i+1} - t)}{t_{i+1} - t_i}\right] \\
&= v_i + \frac{1}{2}(v_i - v_{i+1})\frac{d}{dt}\left[\frac{(t - t_i)(t_{i+1} - t)}{t_{i+1} - t_i}\right] \\
&= v_i + \frac{1}{2}(v_i - v_{i+1})\frac{1}{t_{i+1} - t_i}\left[(t_{i+1} - t) - (t - t_i)\right] \\
&= v_i + \frac{1}{2}(v_i - v_{i+1})\frac{t_{i+1} - t - t + t_i}{t_{i+1} - t_i} \\
&= v_i + \frac{1}{2}(v_i - v_{i+1})\frac{t_{i+1} - t_i - 2(t - t_i)}{t_{i+1} - t_i} \\
&= v_i + \frac{1}{2}(v_i - v_{i+1})\left[1 - \frac{2(t - t_i)}{t_{i+1} - t_i}\right] \\
&= v_i + \frac{1}{2}(v_i - v_{i+1})\left[1 - 2\left(1-\frac{t_{i+1} - t}{t_{i+1} - t_i}\right)\right] \\
&= v_i + \frac{1}{2}(v_i - v_{i+1})\left[1 - 2(1-\alpha_t)\right] \\
&= v_i + \frac{1}{2}(v_i - v_{i+1})(2\alpha_t - 1) \\
\end{align}



The variance function defined in Eq. \eqref{eq:combined_mu_sigma}:

\begin{align}
\sigma(t) &= \sigma_0 \cdot \frac{(t-t_{i})(t_{i+1}-t)}{(t_{i+1}-t_i)}
\end{align}

Let $\Delta t = t_{i+1} - t_i$:

\begin{align}
\sigma(t) &= \sigma_0 \cdot \frac{(t-t_{i})(t_{i+1}-t)}{\Delta t}
\end{align}

We compute the derivative with respect to $t$:

\begin{align}
\sigma'(t) &= \sigma_0 \cdot \frac{d}{dt}\left[\frac{(t-t_{i})(t_{i+1}-t)}{\Delta t}\right] \\
&= \sigma_0 \cdot \frac{1}{\Delta t} \cdot \frac{d}{dt}\left[(t-t_{i})(t_{i+1}-t)\right]
\end{align}

Using the product rule for $(t-t_{i})(t_{i+1}-t)$:

\begin{align}
\frac{d}{dt}\left[(t-t_{i})(t_{i+1}-t)\right] &= (1)(t_{i+1}-t) + (t-t_{i})(-1) \\
&= (t_{i+1}-t) - (t-t_{i}) \\
&= t_{i+1} - t - t + t_{i} \\
&= t_{i+1} + t_{i} - 2t
\end{align}

Therefore:

\begin{align}
\sigma'(t) &= \sigma_0 \cdot \frac{t_{i+1} + t_{i} - 2t}{\Delta t} =\sigma_0\,(2\alpha_t-1), \quad \text{where } \alpha_t=\frac{t_{i+1}-t}{\,t_{i+1}-t_i\,}.
\end{align}

\section{IMMFM Training and Forecasting Algorithms}\label{app:algorithms}

This section provides the detailed procedures for the IMMFM framework. \textbf{Algorithm~\ref{alg:immfm_training}} outlines the complete training process, from sampling trajectories to the stochastic gradient update. \textbf{Algorithm~\ref{alg:immfm_forecasting}} details the autoregressive forecasting method used for inference, which can perform either deterministic (ODE) or stochastic (SDE) simulations. A practical consideration for the training procedure is the use of Multi-Marginal Optimal Transport (MMOT) for constructing the ground-truth trajectories. This potentially expensive MMOT problem can be solved offline as a one-time preprocessing step for datasets where the set of longitudinal observations is fixed. The resulting optimal transport plans can loaded during training, which significantly accelerates the optimization process.

\begin{algorithm}[h]
\caption{IMMFM Training}
\label{alg:immfm_training}
\begin{algorithmic}[1]
\State \textbf{Input:} Training data $\mathcal{D} = \{ \mathbf{z}_n \}_{n=1}^N$, initial variance parameter $\sigma_0$, loss weight $\beta$.
\State \textbf{Initialize networks:} Drift $v_{\theta}$, Score $s_{\theta}$, Diffusion $g_{\theta}$.
\State \textbf{while} training \textbf{do}
\State Sample a mini-batch of full trajectories $\{ (x_{t_0}, \ldots, x_{t_M}) \}$ from $\mathcal{D}$.
\For{each trajectory $\mathbf{z} = (x_{t_0}, \ldots, x_{t_M})$ in the mini-batch}
\State Sample $t \sim \mathcal{U}(t_0, t_M)$.
\State Find index $j$ such that $t \in [t_j, t_{j+1})$.
\State Select segment data: $t_a \leftarrow t_j$, $x_a \leftarrow x_{t_j}$, $t_b \leftarrow t_{j+1}$, $x_b \leftarrow x_{t_{j+1}}$.
\State Set conditioning variable $c \leftarrow x_{t_{j-1}}$ (or zero vector if $j=0$).
\State Define local interp. velocities $v_j \leftarrow \frac{x_b - x_a}{t_b - t_a}$ and $v_{j+1} \leftarrow \frac{x_{t_{j+2}} - x_b}{t_{j+2} - t_b}$ (or $v_j$ if $j=M-1$).
\State Compute $\alpha_t \leftarrow \frac{t_b - t}{t_b - t_a}$.
\State Compute $\mu_t \leftarrow x_a + v_j (t - t_a) + \frac{1}{2}\alpha_t (v_j - v_{j+1})(t - t_a)$.
\State Compute $\mu'_t \leftarrow v_j + \frac{1}{2}(v_j - v_{j+1})(2\alpha_t - 1)$.
\State Sample $x \sim \mathcal{N}(x \mid \mu_t, \sigma(t)^2 I)$, with $\sigma(t) = \sigma_0 (t-t_a)\alpha_t$.
\State Compute target velocity $u^\circ_t(x \mid \mathbf{z}) \leftarrow \frac{\sigma'(t)}{\sigma(t)}(x - \mu_t) + \mu'_t$, with $\sigma'(t) = \sigma_0(2\alpha_t-1)$.
\State Compute target score $\nabla_x \log p_t(x \mid \mathbf{z}) \leftarrow \frac{\mu_t - x}{\sigma(t)^2}$.
\State Compute $\mathcal{L}_{\rm CSDE} \leftarrow \| v_{\theta}(t,x,c) - u^\circ_t(x \mid \mathbf{z}) \|_2^2 + \lambda(t)^2 \| s_{\theta}(t,x,c) - \nabla_x \log p_t(x \mid \mathbf{z}) \|_2^2$.
\State Assemble SDE drift $u_{\theta}(t,x,c) \leftarrow v_{\theta}(t,x,c) + \frac{g_{\theta}(t,x,c)^2}{2} s_{\theta}(t,x,c)$.
\State Predict endpoint $\hat{x}_{t_{b}} \leftarrow x + (t_b - t) u_{\theta}(t,x,c)$.
\State Compute $\mathcal{L}_{\rm uncertainty} \leftarrow \| g_{\theta}(t,x,c)^2 - \| \hat{x}_{t_{b}} - x_b \|_2^2 \|_2^2$.
\State Compute $\mathcal{L}_{\rm IMMFM}(\theta) \leftarrow \mathcal{L}_{\rm CSDE} + \beta \mathcal{L}_{\rm uncertainty}$.
\State Update $\theta$ using $\nabla_{\theta} \mathcal{L}_{\rm IMMFM}(\theta)$.
\EndFor\\
\textbf{end while}
\State \textbf{Output:} Trained networks $v_{\theta}, s_{\theta}, g_{\theta}$.
\end{algorithmic}
\end{algorithm}

\begin{algorithm}[h!]
\caption{IMMFM Forecasting}
\label{alg:immfm_forecasting}
\begin{algorithmic}[1]
\State \textbf{Input:} Trained networks $v_{\theta}, s_{\theta}, g_{\theta}$; observed prefix $(x_{t_0}, \ldots, x_{t_k})$ with $k \ge 1$; forecast end time $t_{end}$; integration step size $\Delta t$; mode $\in$ {'ODE', 'SDE'}.
\State \textbf{Initialize:} Trajectory $\mathcal{T} \leftarrow [x_{t_0}, \ldots, x_{t_k}]$.
\State Set current time $t \leftarrow t_k$.
\State Set current state $x \leftarrow x_{t_k}$.
\State Set conditioning variable $c \leftarrow x_{t_{k-1}}$.
\State Set number of steps $N_{steps} \leftarrow \lfloor(t_{end} - t) / \Delta t\rfloor$.
\For{$i = 0$ to $N_{steps} - 1$}
\If{mode = 'ODE'}
    \State Compute ODE drift $u_t \leftarrow v_{\theta}(t,x,c)$.
    \State Evolve state $x_{new} \leftarrow x + u_t \Delta t$.
\Else{ (mode = 'SDE')}
    \State Sample noise $\mathbf{z} \sim \mathcal{N}(0, I)$.
    \State Compute SDE drift $u_t \leftarrow v_{\theta}(t,x,c) + \frac{g_{\theta}(t,x,c)^2}{2} s_{\theta}(t,x,c)$.
    \State Get SDE diffusion $g_t \leftarrow g_{\theta}(t,x,c)$.
    \State Evolve state $x_{new} \leftarrow x + u_t \Delta t + g_t \sqrt{\Delta t}\mathbf{z}$.
\EndIf
\State Update conditioning variable $c \leftarrow x$.
\State Update current state $x \leftarrow x_{new}$.
\State Update current time $t \leftarrow t + \Delta t$.
\State Append $x$ to $\mathcal{T}$.
\EndFor
\State \textbf{Output:} Complete trajectory $\mathcal{T}$ (observed prefix + forecast).
\end{algorithmic}
\end{algorithm}

\newpage
\section{Datasets}
\label{app:datasets}

\paragraph{S-shape and $\sigma$-shape Gaussian Dataset.}

The S-shaped and $\sigma$-shaped Gaussians both consist of 8 marginal distributions in $\mathbb{R}^2$ at arbitrary timepoints $T = (0, 0.17, 0.29, 0.45, 0.65, 0.71,0.85, 1)$.
We select these two datasets because S-shaped Gaussians involve learning a flow with changing curvature, 
and the $\sigma$-shaped Gaussians have a crossover point for some $x$ where the flow $u_{t_i}(x) = u_{t_j}(x)$ 
and $i \neq j$. 

\textbf{ADNI1 Dataset.} ADNI-1, the inaugural phase of the Alzheimer’s Disease Neuroimaging Initiative \citep{Mueller2005}, launched in October 2004 as a five-year multicenter study, enrolling 317 participants—100 cognitively normal (CN) elderly controls, 117 with amnestic mild cognitive impairment (MCI), and 100 with early Alzheimer’s disease(AD) —across 57 sites in the US and Canada. Participants underwent serial 1.5T structural MRI at approximately six-month intervals. For the sake of simplicity, we use only CN and AD subjects. The dataset is provided with the segmentation mask for the ventricle.

\textbf{Brain Multiple Sclerosis Dataset.} We used longitudinal FLAIR-weighted MRI scans from the Brain MS dataset~\citep{Carass2017}, monitoring \bc{19} patients with multiple sclerosis (MS) over an average of 4.4 time points spanning approximately five years. The Training set included manual delineations by two experts, identifying and segmenting the lesions.

\textbf{Brain Glioblastoma Dataset.} We used longitudinal contrast-enhanced T1-weighted MRI scans from the LUMIERE dataset~\citep{Suter2022}, tracking 91 glioblastoma (GBM) patients who underwent a pre-operative scan followed by repeated post-operative scans over up to five years. This resulted in 795 longitudinal image series, each comprising 2–18 time points. This also comes with segmentation labels for necrosis, contrast enhancement, and edema.

\textbf{Starmen.} The public synthetic Starmen dataset comprises 1\,000 sequences of 10 images each and is commonly used to benchmark longitudinal frameworks~\citep{Bone2018}. In every sequence, the sole temporal change is the raising of the left arm, with each subject’s motion encoded via an affine time parameterization: $t^* = \alpha\,(t - \tau)$, where \(\alpha\) and \(\tau\) are subject-specific parameters. To introduce additional variability, sequences are randomly rotated (uniformly between \(-10^\circ\) and \(10^\circ\)) and translated by up to \(\pm 6.8\) pixels. Of the 1\,000 sequences, 400 are reserved for training, 100 for validation, and the remaining 500 for testing. The ground-truth progression values have a mean of \(-0.12\) and a standard deviation of \(4.25\). Finally, we augment the original dataset by adding two more classes of motion. From only a hand going up motion for each of the splits, we create an equal number of trajectories for two more classes (Hand-downward motion, and static) by reversing the order of the trajectory and replicating the first image 10 times over.

\section{Evaluation Metrics}
\label{app:evaluation_metrics}
We assess the performance of the model using three primary evaluation metrics: image similarity, residual magnitude, and regions of interest (ROI) similarity.

\textbf{Image Similarity: }
Image similarity between the real future image $x_{t_j}$ and the predicted future image $\hat{x}_{t_j}$ is quantified using two widely recognized metrics: Peak Signal-to-Noise Ratio (PSNR) and Structural Similarity Index Measure (SSIM). These metrics are standard for image-to-image tasks, including super-resolution, denoising, and inpainting. PSNR is a logarithmic metric that normalizes the Mean Squared Error (MSE) between two images using their dynamic range. It is defined as:
\begin{equation}
\operatorname{PSNR}(x_a, x_b) = 10 \log_{10} \left( \frac{R^2}{\operatorname{MSE}(x_a, x_b)} \right)
\label{eq:psnr}
\end{equation}
where $R$ is the common dynamic range of the images. The Mean Squared Error (MSE) is calculated as:
\begin{equation}
\operatorname{MSE}(x_a, x_b) = \frac{1}{H \times W} \sum_{h \in H, w \in W} \|x_{a(h,w)} - x_{b(h,w)}\|^2
\label{eq:mse}
\end{equation}

SSIM measures the perceptual similarity between two images, capturing structural changes. The formula is defined as:
\begin{equation}
\operatorname{SSIM}(x_a, x_b) = \frac{\left(2\mu_{x_a} \mu_{x_b} + c_1\right)\left(2\sigma_{x_a x_b} + c_2\right)}{\left(\mu_{x_a}^2 + \mu_{x_b}^2 + c_1\right)\left(\sigma_{x_a}^2 + \sigma_{x_b}^2 + c_2\right)}
\label{eq:ssim}
\end{equation}
where $\mu_{x_a}$ and $\mu_{x_b}$ are the pixel sample means, $\sigma_{x_a}^2$ and $\sigma_{x_b}^2$ are the variances, $\sigma_{x_a x_b}$ is the covariance of $x_a$ and $x_b$, and $c_1 = (0.01R)^2, c_2 = (0.03R)^2$ are constants for numerical stability.

\textbf{Residual Magnitude: }
We also assess the magnitude of residual differences between the predicted and real images using Mean Squared Error (MSE).

\textbf{Region of Interest (ROI) Similarity: }
To accurately capture the ventricle for ADNI and lesion/tumor regions, we use two primary metrics: Dice Similarity Coefficient (DSC) and Hausdorff Distance (HD). These metrics are computed on the binarized atrophy region masks of the real future image $x_{t_j}$ and the predicted future image $\hat{x}_{t_j}$. The DSC and HD are defined as follows respectively:

\begin{equation}
\text{DSC}(X, Y) = \frac{|X \cap Y|}{|X| + |Y|}
\label{dsc}
\end{equation}

\begin{equation}
\text{HD}(X, Y) = \max \left( \sup_{x \in X} d(x, Y), \sup_{y \in Y} d(X, y) \right)
\label{hd}
\end{equation}

\section{Implementation Details} \label{app:implementation_details}


\subsection{Architecture}
\label{app:architecture}
\textbf{Flow Regressor:} The flow regressor network for our proposed model, IMMFM, is built upon the U-ViT \citep{yu2022unified} architecture; we specifically adapt the implementation from \citet{davtyan2023efficient}. The network consists of 14 standard ViT\citep{dosovitskiy2021an} blocks. These are interconnected by 4 long skip connections that link the first 4 blocks to the last 4 blocks. Along each skip connection, feature maps are channel-wise concatenated and subsequently projected to the inner dimension of the ViT blocks. Within each ViT block, Layer Normalization is applied before both the Multihead Self-Attention (MHSA) layer and the subsequent MLP. The inner dimension for all ViT blocks is 512, and 8 heads are used in every MHSA layer. For processing the inputs first, the input image $x_{t_i}$ and conditioning $c={x_{t_{i-1}}}$ are channel-wise concatenated and then linearly projected to the inner dimension of the ViT blocks. As for non-imaging conditioning, such as demographics (e.g., sex, age, etc), we use separate projection blocks to project them to the inner dimension and subsequently add them to the imaging inputs. If no preceding image exists, a standard prior (zero vector) was used. Before passing it to the ViT blocks, we add learned positional embedding and a sinusoidal time embedding \citep{vaswani2017attention} of corresponding time  $t_i$ and $t_{i-1}$ of \bc{image latents} $x_{t_i}$ and $x_{t_i}$. Finally, the network outputs velocity $v_\theta$, score $s_\theta$, and uncertainty $g_\theta$.

\textbf{Auto Encoder:} For encoding the image to the latent space, we use the UNet architecture proposed in \citet{dhariwal2021diffusion}. However, we drop the long skip connection from encoder to decoder to transform it into an autoencoder so that all the information of the input image is contained within a single latent vector. For the Starman dataset, we use a latent dimension of 256, and for all the clinical datasets, we use 4096.

\textbf{Segmentation Network: } For ROI segmentation, we trained three auxiliary image segmentation networks, each tailored to one of the datasets. These networks follow a UNet \citep{isensee2021nnu} architecture.

\subsection{Data Preprocessing and Augmentation}
\label{app:data-preprocessing-aug}
For all three clinical datasets, we register the 3D volumes to have spatial alignment in each trajectory. For this, we use ANTS \citep{Avants2009} to perform Affine followed by Diffeomorphic registration to align each scan towards the first scan in the sequence.

\bc{Following registration, we employed two distinct processing pipelines. For our 2D analysis, we extracted all axial slices containing the primary region of interest (e.g., ventricles in ADNI), a selection guided by the provided ground truth segmentation masks. Each of these 2D intensity slices was then resized to \(256 \times 256\) pixels using cubic interpolation. For our 3D analysis, we used the entire registered volume, resizing each intensity volume to \(128 \times 128 \times 128\) voxels with cubic interpolation. All corresponding segmentation mask volumes were resized using nearest-neighbor interpolation to preserve label integrity.}

The timeline for each trajectory, originally in days and weeks, was scaled to a range between 0 and 1 by dividing by the maximum duration of the respective dataset. For the Starmen dataset, no registration was needed. We employed a two-stage augmentation strategy. During autoencoder pre-training, we used standard image-level augmentations, including flipping, shifting, scaling, and rotation. \bc{However, for the main Flow Matching (FM) training, no image-level augmentation was used. Instead, we performed trajectory augmentation by subsampling. For a given sequence of \(M\) visits, we generated additional training samples by creating all possible contiguous sub-trajectories of shorter lengths, while strictly preserving the temporal order of the images.} 

\bc{We performed a subject-level partition for all datasets to prevent data leakage, ensuring all data from a single subject remained in the same set. For the larger ADNI (317 subjects) and GBM (91 subjects) cohorts, we used a \%70\%10\%/20\% split, resulting in approximately 220/30/67 (train/val/test) subjects for ADNI and 60/8/23 for GBM. For the smaller MS dataset (19 subjects), we performed a 5-fold cross-validation; the results reported correspond to a representative fold with 12 training and 7 test subjects. For each subject, their single 3D volume trajectory was expanded into 20 to 100 2D slice-based trajectories, significantly increasing the number of samples for training.}

\subsection{Training Details}
\label{app:training_details}

\bc{
\textbf{Autoencoder Training.} The first stage of our pipeline involves training an autoencoder to learn a compact latent representation of the 2D images. The architecture is based on the U-Net from \citet{dhariwal2021diffusion}, but with the long skip connections between the encoder and decoder removed to ensure a compressed latent bottleneck. The network incorporates residual layers with convolutions and multi-head self-attention layers. The latent space dimension was set to 4096 for clinical datasets and 256 for the Starmen dataset. The model was trained for 100 epochs using a hybrid loss function formulated as $\mathcal{L} = \text{MSE} + 0.5 \cdot (1 - \text{SSIM})$. Upon completion of training, the autoencoder weights were frozen.}

\bc{
\textbf{Segmentation Network Training.} For downstream tasks requiring biomarker quantification, we trained an auxiliary segmentation network. This network, based on a standard U-Net architecture \citep{isensee2021nnu}, was trained for 100 epochs using a binary cross-entropy loss on the ground truth annotations provided with each dataset. Crucially, the input images for this network were first passed through the complete, pre-trained autoencoder (encoder followed by decoder). This step ensures the segmentation model learns to operate on images that have the same distributional characteristics as our model's generated outputs.}

\textbf{Flow Matching Model Training.} In the second stage, the IMMFM model was trained to learn the progression dynamics directly within the latent space. For each training step, input trajectories were first passed through the frozen encoder to obtain their corresponding latent representations. The IMMFM model was then trained for a longer duration, typically between 350 and 500 epochs. During inference, the trained IMMFM model operates on the latent vectors to produce a forecasted latent state, which is subsequently transformed back into the pixel domain by the pre-trained decoder.
\bc{
Key hyperparameters for all models are detailed in Table~\ref{tab:all_params} and ~\ref{tab:3d_all_params} for 2D and 3D models, respectively. Note that we use 8 ViT blocks for the MS dataset and 14 ViT blocks for the GBM and ADNI datasets.}

\vspace{0.5cm}

\begin{table}[th]
\centering
\caption{\bc{Model Hyperparameters for 2D version}.}
\label{tab:all_params}
\begin{tabular}{lccc}
\toprule
\textbf{Hyperparameter}         & \textbf{Autoencoder}     & \textbf{Segmentation Net} & \textbf{Flow Regressor} \\
\midrule
Model Size                      & $\sim$38M                     & $\sim$3M  & $\sim$40-60M     \\
Input Channels                  & 1                        & 1                         & -        \\
Image Size                      & $256 \times 256$         & $256 \times 256$          & 4096                       \\
Architecture                   & CNN + Attention            & U-Net                    & ViT-based               \\
Transformer Blocks              & -                        & -                         & 8-14 ViT Blocks           \\
Skip Connections                & -                        & Yes                       & Yes \\
ViT Inner Dimension             & -                        & -                         & 512                     \\
Channels                        &  64                      & 16                         & -            \\
Channel Multiple                & 2,4,4,4,4,4,4          & 1,2,4,8,16                   & -    \\
Residual Blocks per Down Block  & 1                      & 2                           & -              \\
Channels / Attention Heads      & 8                        & -                         & 8                       \\
Attention Resolution            & 64,32,16,8,4,2,1         & -                         & -            \\
Dropout                         & 0.0                      & 0.0                       & 0.0                     \\
Batch Size                      & 24                        & 24                        & 24                      \\
Epochs                          & 100                       & 150                       & 350--500                \\
Warmup Epochs                   & 25                       & 25                        & 25                       \\
Learning Rate                   & $1 \times 10^{-3}$       & $1 \times 10^{-4}$        & $1 \times 10^{-4}$   \\ 
\bottomrule
\end{tabular}
\end{table}

\vspace{0.5cm}

\begin{table}[th]
\centering
\caption{\bc{Model Hyperparameters for 3D version.}}
\label{tab:3d_all_params}
\begin{tabular}{lccc}
\toprule
\textbf{Hyperparameter}         & \textbf{Autoencoder}     & \textbf{Segmentation Net} & \textbf{Flow Regressor} \\
\midrule
Model Size                      & $\sim$45M          & $\sim$5M             & $\sim$75M          \\
Input Channels                  & 1                           & 1                             & -                           \\
Input Volume Size               & $128 \times 128 \times 128$  & $128 \times 128 \times 128$  & 131,072                     \\
Architecture                   & 3D CNN + Attention          & 3D U-Net                       & ViT-based                \\
Transformer Blocks              & -                           & -                             & 14 ViT Blocks               \\
Skip Connections                & -                           & Yes                           & Yes                         \\
ViT Inner Dimension             & -                           & -                             & 512                         \\
Channels                        &  32                        & 16                         & -            \\
Channel Multiple                & 2,4,4,8,8                  & 1,2,4,8,16                   & -    \\
Residual Blocks per Down Block  & 1                           & 2                           & -              \\

Channels / Attention Heads      & 8                           & -                             & 8                           \\
Attention Resolution            & 64,32,16                    & -                             & -                \\
Dropout                         & 0.0                         & 0.0                           & 0.0                         \\
Batch Size                      & 4                           & 4                             & 2                           \\
Epochs                          & 100                         & 150                           & 350--500                    \\
Warmup Epochs                   & 25                          & 25                            & 25                          \\
Learning Rate                   & $2 \times 10^{-4}$          & $2 \times 10^{-4}$            & $1 \times 10^{-5}$          \\
\bottomrule
\end{tabular}
\end{table}

\subsection{Computing Infrastructure and Cost}

All experiments were performed on Snellius, the Dutch national supercomputer. Each training job was allocated a node equipped with one NVIDIA A100 GPU (with 40GB VRAM) and 8 CPU cores. 
Typical training durations for our primary models on this configuration were as follows:
\begin{itemize}
    \item Autoencoder: 12--16 hours.
    \item Segmentation Network: 6--8 hours.
    \item IMMFM: 12--18 hours.
\end{itemize}

In comparison to our proposed models, some baseline methods exhibited greater computational demands. Notably, the ImageFlowNet baseline consistently required a significantly longer training period, taking approximately 2x longer when executed on similar hardware. The total computational resources utilized for developing our models, conducting all experiments, and performing baseline comparisons in this study amounted to approximately 3000 GPU hours. Our implementation primarily relies on PyTorch. \bc{ Note that both 2D and 3D experiments took similar time due to a reduction in dataset samples when treated as a 3D volume.}

\section{Additional Experimental Results}
\label{app:additional_results}

\subsection{Ablation Study of Proposed Components}
\label{app:ablation-study}
\bc{
To evaluate our methodological contributions, we conducted an ablation study to isolate and quantify the impact of each component of our proposed IMMFM framework. Table~\ref{tab:ablation-study} shows the performance gains from the following components: the piecewise quadratic conditional path, the data-driven diffusion coefficient, and conditioning on the previous frame. Below, we describe the experimental setup for each ablation:
\paragraph{Previous-Frame Conditioning.} To assess the importance of immediate temporal context, we trained our simplest model variant, O-IMMFM, without conditioning on the latent representation of the preceding frame. All other aspects of the experimental setup, including architecture and training hyperparameters, remained unchanged.
\paragraph{Piecewise Quadratic Path.} To measure the benefit of incorporating second-order temporal dynamics, we compared our ODE-based model, O-IMMFM (using the proposed quadratic path,) against a baseline with a conventional linear path that connects consecutive marginals with straight lines. This isolates the performance gain attributable to our novel quadratic path construction.
\paragraph{Learned Diffusion Coefficient.} To quantify the effect of a learned diffusion term, we compared the full SDE-based model SU-IMMFM against S-IMMFM, which uses a fixed, predefined diffusion schedule. Both variants share identical architectures, isolating the impact of using a data-driven approach to modeling stochasticity.}

\bc{The results of our ablation study highlight the value of each proposed component. Introducing the quadratic path yields the most substantial improvements, increasing Dice Score by up to 3.7\% and PSNR by over 2.0 dB. Similarly, incorporating a learned diffusion coefficient consistently improves performance on the most challenging GBM dataset, notably reducing Hausdorff Distance by over 6.2 pixels and increasing Dice Score by 1.5\%. Finally, conditioning on the previous frame improves performance as well, boosting the Dice Score by up to 2.1\% and PSNR by over 1.0 dB.}

\begin{table}[H]
  \centering
  \caption{\bc{Ablation study results quantifying the contribution of each model component. The table reports the change in performance metrics (averaged over 3 seeds) from our ablation study. Each column represents the performance gain from a specific component.}}
  \label{tab:ablation-study}
  \begin{tabular}{llccc}
    \multicolumn{5}{c}{} \\
    \toprule
    Dataset & Metric & \makecell{Prev. frame conditioning} & \makecell{Quadratic path} & \makecell{Data diffusion} \\
    \midrule
    \multirow{5}{*}{ADNI} 
      & PSNR $\uparrow$ &   1.089 & 1.441  & 0.084 \\
      & SSIM $\uparrow$ &    0.021 & 0.012  & 0.001 \\
      & MSE  $\downarrow$ &  0.001 & $>$0.001  & $>$0.001 \\
      & DSC  $\uparrow$   &  0.021 & 0.027  & 0.002 \\
      & HD   $\downarrow$ & 1.198 & 2.16 & 0.671 \\
    \midrule
    \multirow{5}{*}{Brain MS} 
      & PSNR $\uparrow$ &  0.940    & 2.098 & 0.146 \\
      & SSIM $\uparrow$ &  0.032    & 0.015 & 0.000 \\
      & MSE  $\downarrow$ & 0.000   & $>$0.001 & $>$0.001 \\
      & DSC  $\uparrow$  & 0.007    & 0.030 & 0.004 \\
      & HD   $\downarrow$ &0.140  & 0.490 & 0.140 \\
    \midrule
    \multirow{5}{*}{Brain GBM} 
      & PSNR $\uparrow$   & 1.092     & 1.696     & 0.467 \\
      & SSIM $\uparrow$   & 0.013     & 0.025     & 0.012 \\
      & MSE  $\downarrow$ & 0.001    & $>$0.001   & $>$0.001 \\
      & DSC  $\uparrow$   & 0.015     & 0.037     & 0.015 \\
      & HD   $\downarrow$ & 1.030    & 1.523 & 6.290 \\
    \bottomrule
  \end{tabular}
\end{table}

\bc{While global metrics such as SSIM and MSE show minimal changes (e.g., SSIM improves by only 0.01–0.02), this is expected, as they are often saturated due to the already high image reconstruction quality and are less sensitive to localized changes in regions of interest (ROIs). For example, small volumetric changes in structures like the ventricles affect only a small fraction of the image, having negligible influence on overall MSE or SSIM. Since our primary goal is to model clinically relevant regional evolution, improvements in ROI-specific metrics such as Dice and Hausdorff Distance provide stronger evidence of our model's effectiveness.}

\subsection{Analysis of Temporal Generalization}
\label{app:temporal-gen}
\begin{figure}[h]
    \centering
    \includegraphics[width=0.98\linewidth]{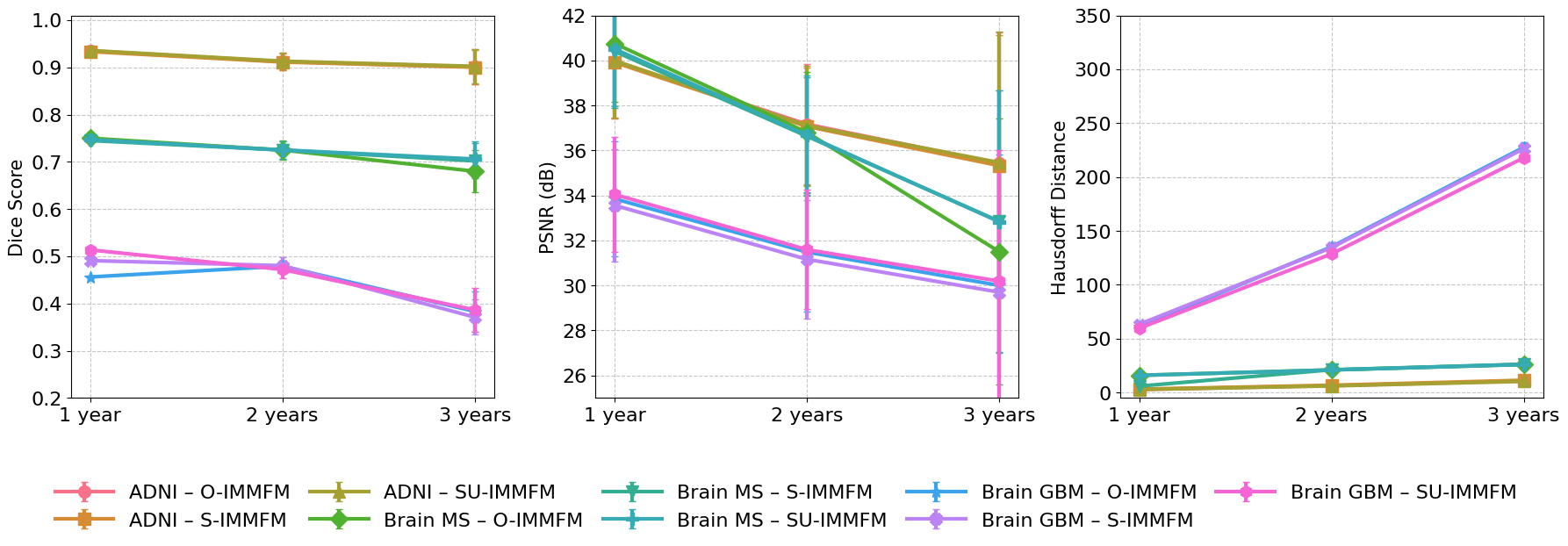}
    \caption{\bc{Performance for increasing forecast horizon for ADNI, MS, and GBM datasets. The error bar represents the average inter-subject variability of the datasets.}}
    \label{fig:time_vs_performance}
\end{figure}

\bc{An important aspect of any forecasting model is understanding how its predictive accuracy changes as the forecast horizon increases. To assess the reliability of our model for both short-term and long-term performance, we evaluate its key metric at various future time points. Fig. \ref{fig:time_vs_performance} illustrates this temporal generalization by plotting Dice score, PSNR, and Hausdorff against the increasing prediction interval, quantifying the expected degradation in accuracy as the model predicts further into the future.}

\bc{The results show an expected degradation in performance as the forecast horizon extends, with Dice Score and PSNR decreasing while Hausdorff Distance increases. The severity of this degradation differs across datasets; on the ADNI dataset, the Dice Score degrades by approximately $\sim$3\% over three years, while the Brain MS dataset sees a similar drop of $\sim$4\%. However, the Brain GBM dataset exhibits a much sharper decline in performance, with up to $\sim$ 13\%. This is likely due to the difficult and sporadic nature of glioblastoma progression, which is highly unpredictable and challenging to model accurately, particularly with limited training data.}

\subsection{Generalization to Volumetric Data}
\label{app:3d-experiment}
\bc{To demonstrate the generalization capability of our framework to volumetric data, we conducted our primary experiments on the full 3D clinical datasets. The experimental setup, including dataset partitions and training protocols, remained identical to the 2D experiments. The core architecture of our flow regressor was also unchanged. To handle the volumetric data, we modified the autoencoder and the auxiliary segmentation network by replacing their 2D convolutional layers with their 3D counterparts. Following spatial alignment, the full 3D volumes were directly processed by these models to create and analyze the latent space trajectories. All the evaluation metrics that are defined in \ref{app:evaluation_metrics} were generalized to work on 3D data for computing the performance. More details can be found about data pre-processing and training in \ref{app:data-preprocessing-aug} and \ref{app:training_details}.}

\begin{table}[h!]
\centering
\caption{Trajectory forecasting performance of the 3D models. Results are averaged over three runs and \emph{all available timepoints}. Values are presented as Mean, with the subscript in \textcolor{green}{green} indicating inter-subject variability and the superscript in \textcolor{purple}{purple} indicating inter-model variability.}
\label{tab:3d-result}
\setlength{\tabcolsep}{15pt}
\begin{tabular}{llccc}
\multicolumn{5}{c}{} \\
\toprule
Dataset & Metric & \makecell{O-IMMFM} & \makecell{S-IMMFM} & \makecell{SU-IMMFM} \\
\midrule
\multirow{5}{*}{ADNI} 
 & PSNR $\uparrow$ & $41.07_{\textcolor{green}{\pm 2.84}}^{\textcolor{purple}{\pm 0.11}}$ & $40.98_{\textcolor{green}{\pm 2.80}}^{\textcolor{purple}{\pm 0.11}}$ & $41.15_{\textcolor{green}{\pm 2.84}}^{\textcolor{purple}{\pm 0.20}}$ \\[4pt]
 & SSIM $\uparrow$ & $0.980_{\textcolor{green}{\pm 0.039}}^{\textcolor{purple}{\pm 0.001}}$ & $0.979_{\textcolor{green}{\pm 0.039}}^{\textcolor{purple}{\pm 0.002}}$ & $0.980_{\textcolor{green}{\pm 0.039}}^{\textcolor{purple}{\pm 0.001}}$ \\[4pt]
 & MSE $\downarrow$ & $0.001_{\textcolor{green}{\pm 0.004}}^{\textcolor{purple}{\pm 0.001}}$ & $0.001_{\textcolor{green}{\pm 0.004}}^{\textcolor{purple}{\pm > 0.001}}$ & $0.001_{\textcolor{green}{\pm 0.004}}^{\textcolor{purple}{\pm 0.000}}$ \\[4pt]
 & DSC $\uparrow$ & $0.938_{\textcolor{green}{\pm 0.077}}^{\textcolor{purple}{\pm 0.009}}$ & $0.942_{\textcolor{green}{\pm 0.081}}^{\textcolor{purple}{\pm 0.004}}$ & $0.947_{\textcolor{green}{\pm 0.070}}^{\textcolor{purple}{\pm 0.005}}$ \\[4pt]
 & HD $\downarrow$ & $2.889_{\textcolor{green}{\pm 2.687}}^{\textcolor{purple}{\pm 0.054}}$ & $3.021_{\textcolor{green}{\pm 2.777}}^{\textcolor{purple}{\pm 0.231}}$ & $2.877_{\textcolor{green}{\pm 2.692}}^{\textcolor{purple}{\pm 0.052}}$ \\
\midrule
\multirow{5}{*}{Brain GBM} 
 & PSNR $\uparrow$ & $34.38_{\textcolor{green}{\pm 5.13}}^{\textcolor{purple}{\pm 0.31}}$
                   & $34.63_{\textcolor{green}{\pm 4.90}}^{\textcolor{purple}{\pm 0.26}}$
                   & $34.15_{\textcolor{green}{\pm 4.03}}^{\textcolor{purple}{\pm 0.61}}$ \\[4pt]
                   
 & SSIM $\uparrow$ & $0.923_{\textcolor{green}{\pm 0.181}}^{\textcolor{purple}{\pm 0.001}}$
                   & $0.923_{\textcolor{green}{\pm 0.167}}^{\textcolor{purple}{\pm 0.002}}$
                   & $0.935_{\textcolor{green}{\pm 0.193}}^{\textcolor{purple}{\pm 0.001}}$ \\[4pt]
 & MSE  $\downarrow$ & $0.001_{\textcolor{green}{\pm 0.001}}^{\textcolor{purple}{\pm >0.001}}$
                     & $0.001_{\textcolor{green}{\pm 0.001}}^{\textcolor{purple}{\pm >0.001}}$
                     & $0.001_{\textcolor{green}{\pm 0.001}}^{\textcolor{purple}{\pm >0.001}}$ \\[4pt]
 & DSC  $\uparrow$ & $0.478_{\textcolor{green}{\pm 0.152}}^{\textcolor{purple}{\pm 0.009}}$
                   & $0.480_{\textcolor{green}{\pm 0.161}}^{\textcolor{purple}{\pm 0.004}}$
                   & $0.489_{\textcolor{green}{\pm 0.150}}^{\textcolor{purple}{\pm 0.005}}$ \\[4pt]
 & HD   $\downarrow$ & $127.61_{\textcolor{green}{\pm 28.19}}^{\textcolor{purple}{\pm 2.054}}$
                     & $124.40_{\textcolor{green}{\pm 29.91}}^{\textcolor{purple}{\pm 1.231}}$
                     & $121.11_{\textcolor{green}{\pm 27.89}}^{\textcolor{purple}{\pm 1.520}}$ \\

\bottomrule
\end{tabular}
\end{table}

Our results in Table~\ref{tab:3d-result} show that the 3D models achieve a notable improvement in forecasting accuracy when compared to their 2D counterparts in Table~\ref{tab:forecasting}. For the ADNI dataset, the SU-IMMFM variant shows marked improvements across the board: the Dice Similarity Coefficient (DSC) increases from 0.920 to 0.947 (an \emph{improvement of 2.9\%}), the PSNR rises from 37.52 to 41.15, and the Hausdorff Distance drops significantly from 6.50 to 2.88.

A similar trend is observed for the Brain GBM dataset, where the SU-IMMFM model improves the DSC from 0.460 to 0.489, a more pronounced \emph{increase of 6.3\%}, and reduces the Hausdorff Distance from 135.08 to 121.11. Across both datasets, not only did the average performance improve, but the inter-subject variability (i.e., the standard deviation) also decreased across most metrics, indicating more consistent predictions.

This performance gain can be attributed to the 3D autoencoder's ability to leverage inter-slice spatial context. By processing the entire volume, it learns a richer latent representation that encodes the full 3D anatomical structure. This is crucial for accurately modeling volumetric, disease-related changes, such as ventricular enlargement in ADNI or the complex, infiltrative growth patterns of tumors in GBM, rather than treating them as disconnected 2D area changes. It should be noted that the Brain MS dataset was excluded from the 3D experiments, as its limited size was insufficient for effectively training the higher-capacity 3D models.

\begin{figure}[t!]
    \centering
    \includegraphics[width=0.95\linewidth]{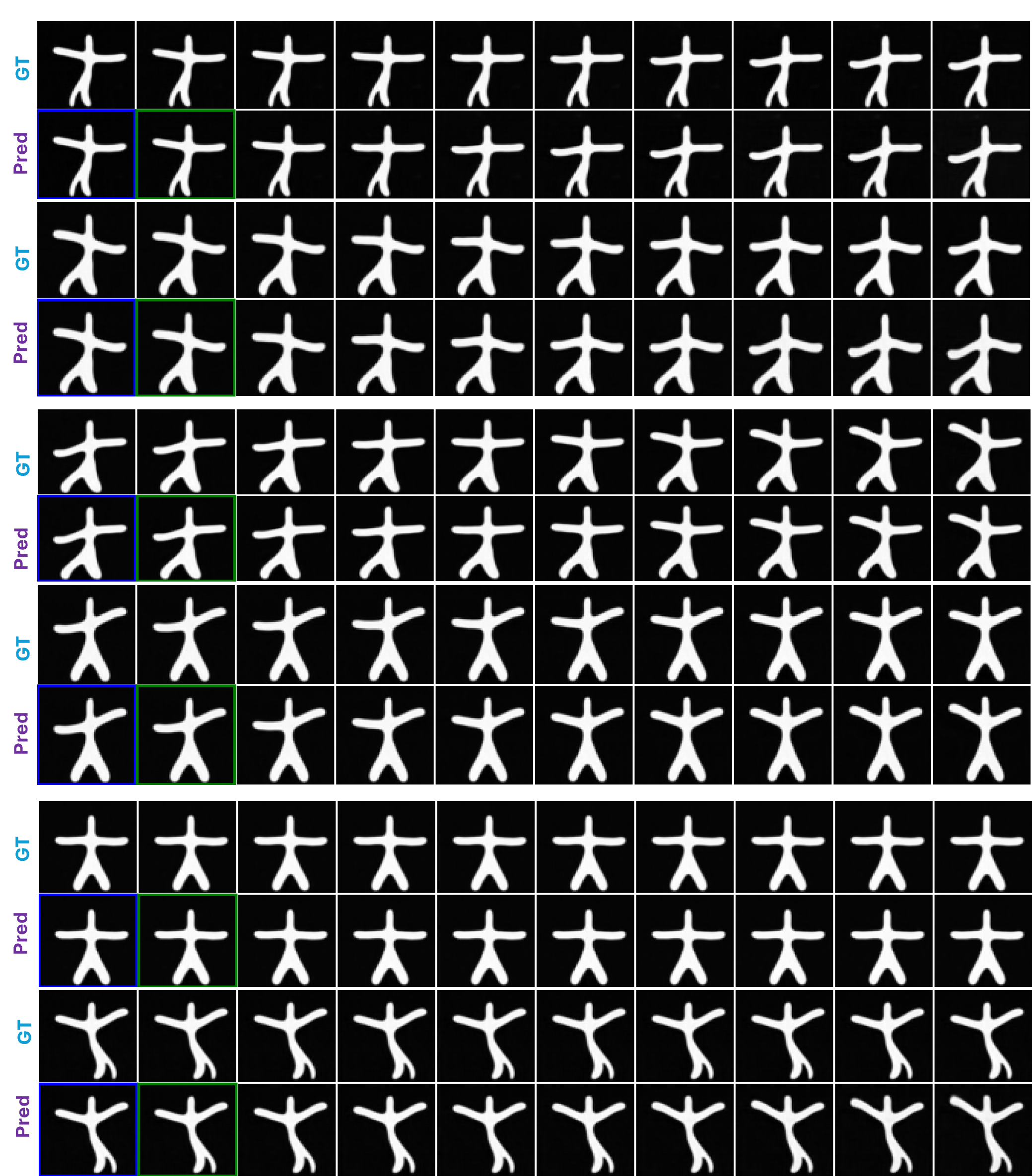}
    \caption{Trajectory simulation on Starmen dataset. The conditioning frame is marked with green, and the reference starting frame is marked with blue. From top to bottom blocks: \emph{Hand-downward motion, Hand-upward motion, Static}.}
    \label{fig:starman_large}
\end{figure}
\section{Additional Methodological Details}
\label{app:additional_method}

\subsection{ADNI Dataset}

\subsubsection{Measuring Overlap}\label{app:ovl}
To quantitatively assess the separation between our AD and CN populations, we fit Gaussians to the ventricular area estimates and measure the overlap between their respective distributions using the Overlap Coefficient (OVL) \citep{inman1989}. For two Gaussian distributions with means $\mu_1$, $\mu_2$ and standard deviations $\sigma_1$, $\sigma_2$, when the variances are unequal, the calculation must account for the two intersection points where the probability density functions meet. The intersection points $(c_1, c_2)$ are determined by:

\begin{figure}[t]
    \centering
    \includegraphics[width=0.98\linewidth]{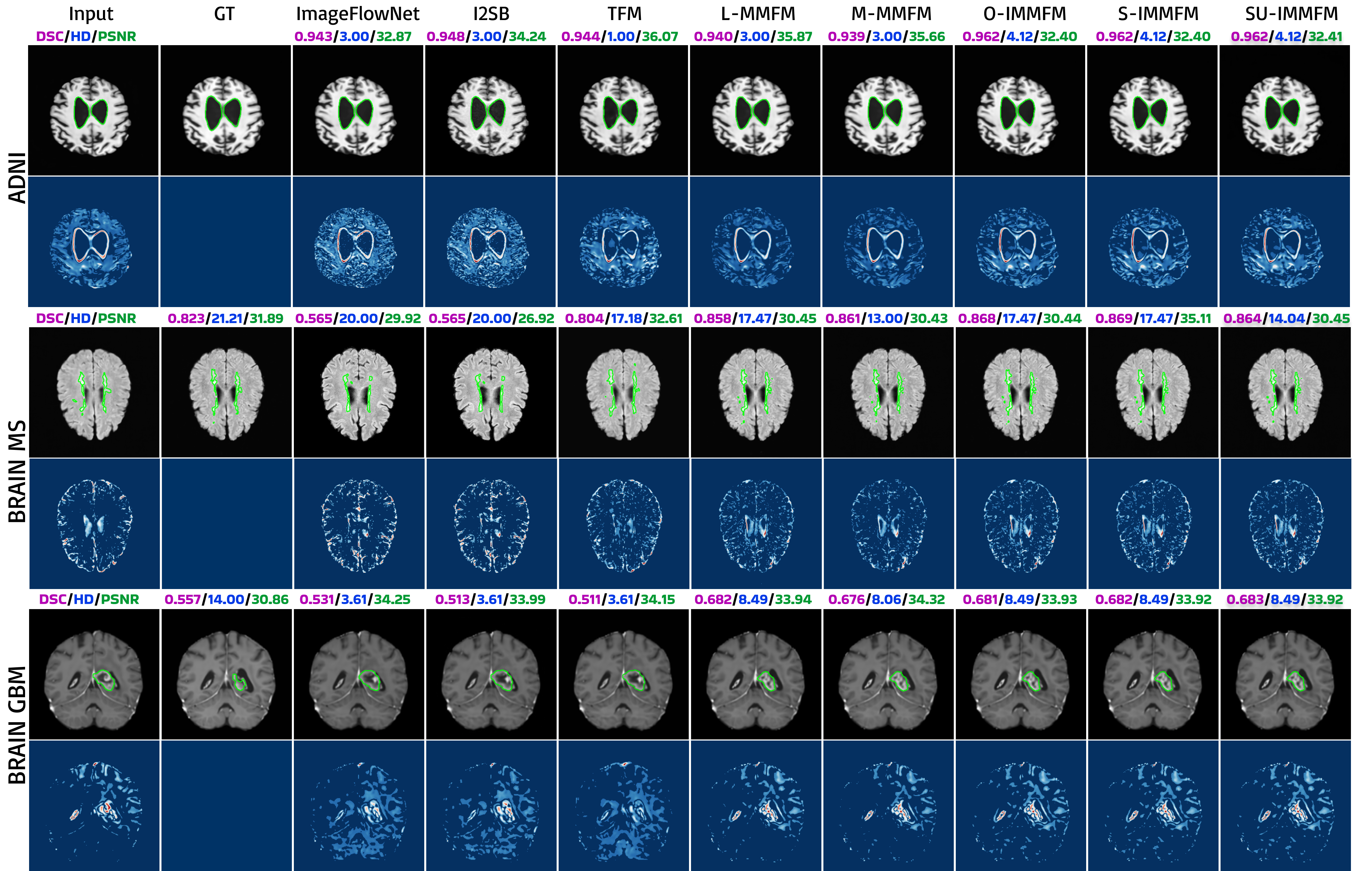}
    \caption{\bc{Additional Visual comparison of forecasting results on the ADNI ($\sim$3yr), MS ($\sim$3yr), and GBM ($\sim$1yr) datasets. For each dataset, the first row displays our model's forecasted image. The second row shows the corresponding pixel-wise difference map between our forecast and the ground truth.}}
    \label{fig:all_traj_appendix}
\end{figure}

\begin{equation}
(c_1, c_2) = \frac{\mu_1\sigma_2^2 - \mu_2\sigma_1^2 \pm \sigma_1\sigma_2\sqrt{(\mu_1-\mu_2)^2 + (\sigma_2^2-\sigma_1^2)\ln(\sigma_2^2/\sigma_1^2)}}{\sigma_2^2-\sigma_1^2}
\end{equation}

The overlap coefficient is then calculated as:

\begin{equation}
\text{OVL} = [1-F_1(c_1)+F_2(c_1)]-[F_2(c_2)-F_1(c_2)]
\end{equation}

where $F_1$ and $F_2$ are the cumulative distribution functions of the respective Gaussian distributions. When both distributions have equal variance, the OVL simplifies to $2\Phi(-|\mu_1-\mu_2|/\sqrt{2\sigma^2})$, where $\Phi$ is the standard normal CDF. This value ranges from 0 (completely separated distributions) to 1 (identical distributions), providing an intuitive measure of classification difficulty \citep{reiser1999}.

\subsubsection{Classification Methodology}\label{app:classification_details}

We performed a binary classification to differentiate Alzheimer's Disease (AD) from Cognitively Normal (CN) subjects, leveraging the normalized ventricular area as a key biomarker. The distinct distributional characteristics of this biomarker between AD and CN populations, which can be analyzed by fitting Gaussian models, motivate its use for this classification task.

The specific feature utilized for classification is the normalized ventricular area, evaluated for each subject in the main test set at two distinct timepoints:
\begin{itemize}
    \item An early observed timepoint, $t_{\text{second}}$ (corresponding to their second clinical visit).
    \item A future timepoint, $t_{\text{last}}$ (normalized time $t=1$, approximately 36 months from baseline), with the ventricular area forecasted from our IMMFM model's predictions.
\end{itemize}

For this specific classification experiment, the main test set was further randomly partitioned to create internal ``threshold-training'' and ``threshold-evaluation'' subsets. To assess the sensitivity of our classification results to the size of these internal partitions, we explored several split ratios. Specifically, we used proportions of $25\%/75\%, 50\%/50\%,$ and $75\%/25\% $ for allocating main test set subjects to the threshold-training versus threshold-evaluation subsets, respectively. For each of these configurations, this partitioning ensured that the determination of the optimal classification threshold and its subsequent performance assessment were conducted on entirely separate (non-overlapping)  subsets.

\begin{table}[ht]
\label{app:tab_adni_classification}
\centering
\caption{Classification Results for ADNI on test set with varying split-ratio}
\begin{tabular}{ccccc}
\hline
\textbf{Train/Test} & \textbf{Second Timepoint} & \textbf{Last Timepoint*}&\textbf{Acc. Gain}& \\ \hline 
50/50  \%  &  67.5 \%   & 75.1\%      & 8.6\%               \\ \hline
75/25 \%    & 71.7\%     & 80.8\%            & 9.1\%          \\
\bottomrule
Average          & 69.6\%            &  78.0 &    8.4\%     \\ \hline
\end{tabular}

\end{table}

While the analysis of distributional overlap (as detailed in Appendix~\ref{app:ovl}) involves identifying intersection points of fitted Gaussian distributions to understand theoretical separability, for the practical task of classifying individual subjects, we determined the decision threshold empirically to optimize predictive performance. 

Using data solely from the ``threshold-training'' subset, an optimal decision threshold for the normalized ventricular area was identified. This was achieved by employing Receiver Operating Characteristic (ROC) curve analysis. The threshold selected was the one that maximized the accuracy in distinguishing AD from CN subjects. The classification threshold learned from the ``threshold-training'' subset was then applied to the ``threshold-evaluation'' subset to assign AD or CN labels to its subjects.

The detailed classification outcomes for different split ratios are presented in Table~\ref{app:tab_adni_classification}.

\begin{figure}[t!]
\includegraphics[width=0.98\linewidth]{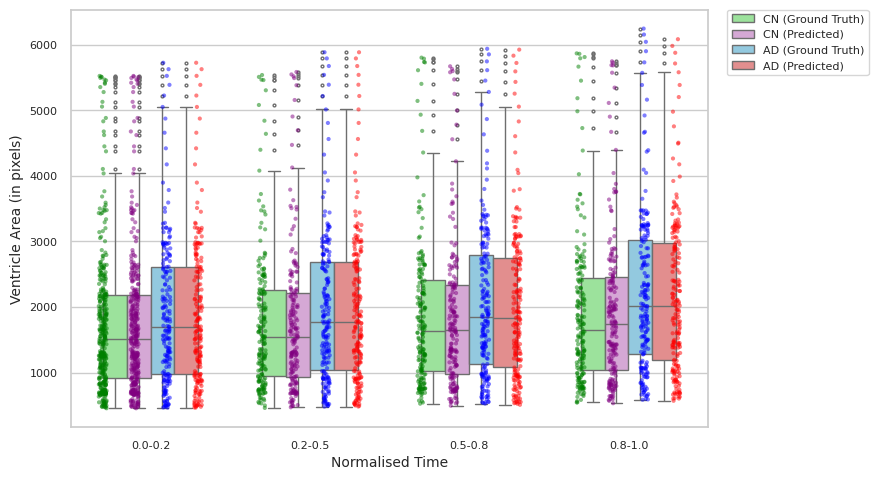}
            \caption{Distribution of ventricular region over time for Alzheimer's (AD) and Cognitively Normal (CN), binned into four discrete time points.}
            \label{fig:ad_cn_dist_full}
                \label{fig:enter-label}
    \end{figure}

\section{Limitations and Future Directions}
\label{app:limitation_future_work}
While IMMFM demonstrates robust performance, its accuracy can be influenced by severe, systematic artifacts in input data, and its predictive scope is shaped by the diversity of trajectories within the training set. These considerations motivate several methodological extensions to enhance the framework's power and versatility.

A primary direction is to learn more informative latent representations. This can be achieved by developing \emph{temporally aware autoencoders}, which move beyond processing snapshots independently and instead employ ordering-aware training objectives or architectural priors to ensure latent space continuity and coherence \citep{yang2023self, blattmann2023align}. Such representations would provide a stronger foundation for several advanced applications. One is enriching the dynamics via \emph{multi-modal conditioning}, allowing the model to integrate heterogeneous data like static covariates or external signals to learn more disentangled and explanatory trajectories \citep{shaik2024survey, wu2024q}. Another is extending the framework's generative capabilities towards \emph{causal and counterfactual reasoning}. By integrating principles of causal representation learning, the model could simulate trajectories under hypothetical interventions, transforming it from a prognostic tool into a system for decision support \citep{von2024causal}.

Finally, to improve plausibility and generalization in data-scarce settings, the model can be fortified with \emph{domain-specific knowledge}. Integrating frameworks from scientific machine learning, such as Physics-Informed Neural Networks (PINNs), can constrain the learned dynamics to adhere to known governing equations \citep{qian2025physics, cuomo2023scientific}. 
\end{document}